\documentclass[10pt,twocolumn,letterpaper]{article}

\usepackage[pagenumbers]{cvpr} 

\usepackage{graphicx}
\usepackage{amsmath}
\usepackage{amssymb}
\usepackage{booktabs}
\usepackage[table,xcdraw]{xcolor}

\usepackage[pagebackref,breaklinks,colorlinks]{hyperref}

\usepackage[capitalize]{cleveref}
\crefname{section}{Sec.}{Secs.}
\Crefname{section}{Section}{Sections}
\Crefname{table}{Table}{Tables}
\crefname{table}{Tab.}{Tabs.}

\def\cvprPaperID{*****} 
\def\confName{CVPR}
\def\confYear{2023}

\begin{document}

\title{wildNeRF: Complete view synthesis of in-the-wild dynamic scenes captured using sparse monocular data}

\author{Shuja Khalid\\
University of Toronto\\
27 King's College Cir, Toronto, ON.\\
{\tt\small skhalid@cs.toronto.edu}
\and
Frank Rudzicz\\
University of Toronto \\
27 King's College Cir, Toronto, ON.\\
{\tt\small frank@cs.toronto.edu}
}

\twocolumn[{%
\renewcommand\twocolumn[1][]{#1}%
\maketitle
\begin{center}
    \centering
    \captionsetup{type=figure}
    \includegraphics[width=0.8\linewidth]{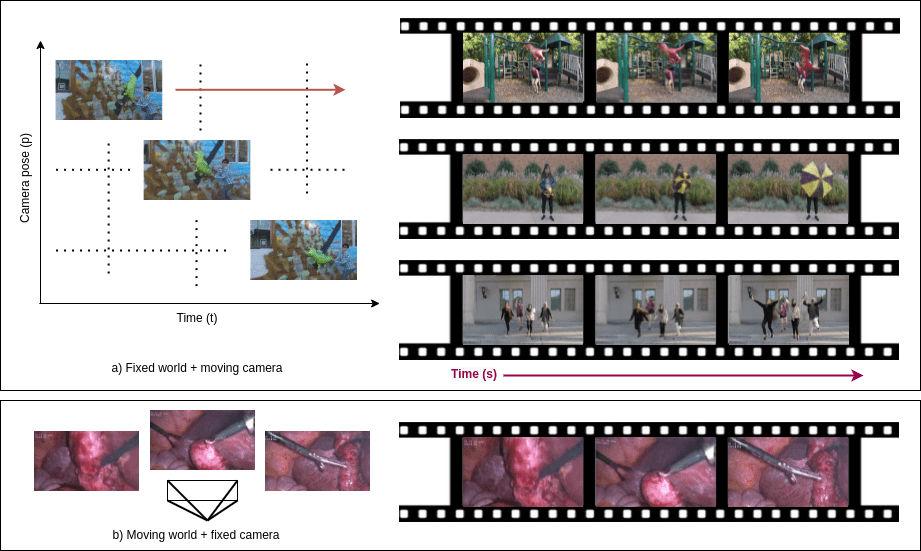}
    \captionof{figure}{Qualitative and quantitative results on two scene capturing approaches for easier adoption of neural view synthesis and motion tracking of animate objects. \textit{a)} assumes a static scene in which the camera pose is changed to \textit{simulate} motion. COLMAP \cite{schonberger2016structureCOLMAP} is then used to extract pose information for in-the-wild scenes. \textit{b)} assumes a fixed camera where the world is in motion, relative to the camera. This setup is common in surgical laparoscopic videos and the pose information can be extracted using COLMAP. We focus on novel view synthesis where the camera pose is fixed and the model synthesizes views at different times.}
   \label{fig:main}
\end{center}%
}]


\begin{abstract}
We present a novel neural radiance model that is trainable in a self-supervised manner for novel-view synthesis of dynamic unstructured scenes. Our end-to-end trainable algorithm learns highly complex, real-world static scenes within seconds and dynamic scenes with both rigid and non-rigid motion within minutes. By differentiating between static and motion-centric pixels, we create high-quality representations from a sparse set of images. We perform extensive qualitative and quantitative evaluation on existing benchmarks and set the state-of-the-art on performance measures on the challenging NVIDIA Dynamic Scenes Dataset. Additionally, we evaluate our model performance on challenging real-world datasets such as Cholec80 and SurgicalActions160.
\end{abstract}

\section{Introduction}
\label{sec:intro}

\begin{figure}
  \centering
   \includegraphics[width=0.8\linewidth]{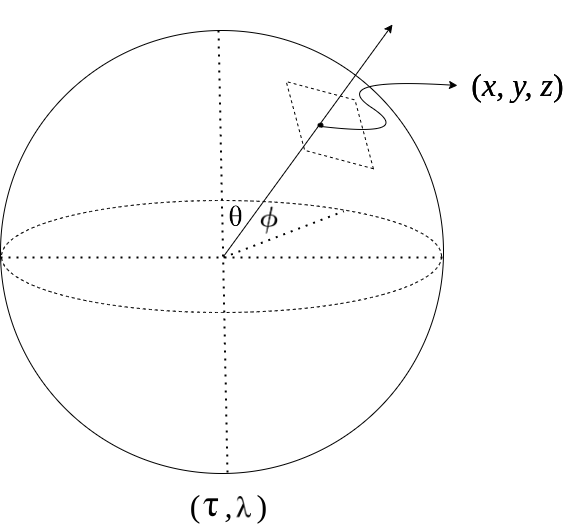}
   \caption{A generalized representation of the Plenoptic function. We constrain the function by considering only pixel coordinates $(x, y, z)$, camera parameters $(\theta, \phi)$, and time $(\tau)$, and assuming a constant wavelength $(\lambda)$.}
   \label{fig:plenoptic-function}
\end{figure}

Visual scene understanding is a long-standing problem in computer vision. The ability of machine learning models to identify, cluster, classify, and track pixels within and across frames has seen tremendous growth over the last decade. Recently, novel-view synthesis and novel-time synthesis have gained significant traction because of neural radiance-based methods \cite{mildenhall2021nerf, park2021nerfies, martin2021nerf, attal2021torfnerf, park2021hypernerf} that combine deep learning approaches and classical vision modelling. This allows for fine-grained reconstructions in 3D space. These models take a sparse set of input frames as input, from which camera pose information is extracted \cite{schonberger2016structureCOLMAP}. A complete scene is generated by modelling a 7D scene representation $f(x, y, z, \theta, \phi, \lambda, \tau)$ where $x, y, z$ are spatial locations and $\theta, \phi$ are camera pose parameters. This generalized representation is known as the \textit{Plenoptic function} as it describes all image information visible from a particular viewing position \cite{bergen1991plenoptic}, for different wavelengths $\lambda$ and time $\tau$. This function may be constrained to a 6D representation by assuming a constant wavelength, $\lambda$. This representation has various real-world applications, including depth estimation \cite{depthchang2019deep, depthguizilini20203d}, 3D tracking \cite{zhou2020tracking, tracklepetit2005monocular}, and video stabilization \cite{liu2012videostab, yang2009robuststab}. We illustrate this generalized formulation in Figure \ref{fig:plenoptic-function}.  

Despite significant advances, existing model architectures are still limited. $1.$ They are extremely computationally intensive and can take days to create believable representations of very small scenes \cite{martin2021nerf, park2021nerfies, park2021hypernerf}. Various works \cite{muller2022instantspeed, reiser2021kilonerfspeed, yu2021plenoctreesspeed, fridovich2022plenoxelsspeed} leverage inefficiencies in neural-radiance architectures and incorporate massively parallel algorithms to speed-up reconstruction times. $2.$ The problem is extremely under constrained; e.g., learning a 6D representation from, in some cases, a set of 12 sparse inputs (NVIDIA dynamic scenes dataset \cite{yoon2020novel}) leads to significant artifacts along the time axis. $3.$ Traditional NeRF-based \cite{martin2021nerf} reconstructions use sparse inputs captured in a more holistic manner to allow for easier generalization.

Our contributions are as follows: $1.$ We showcase the efficacy of our model using the publicly available \textit{NVIDIA dynamic scenes dataset} and extend our model to real-world datasets such as the \textit{SurgicalActions160} \cite{sa160} and \textit{Cholec80} \cite{twinanda2016endonetcholec80} datasets. $2.$ To the best of our knowledge, this is the first paper to incorporate CUDA-accelerated training schemes for monocular dynamic scenes, thereby increasing the applicability of such models to applications. $3.$ We extend state-of-the-art techniques by Li \textit{et al} \cite{li2021neuralnsff} and Gao \textit{et al} \cite{gao2021dynamicnerf} to successfully model both rigid and non-rigid motion. $4.$ We apply this approach to extremely challenging surgical datasets for the purpose of 3D instrument-tracking and video stabilization, the successful implementation of which would open the door to exciting new advances in surgical skill evaluation \cite{khalid2020system, khalid2022or}. 



\section{Related Work}
\label{sec:rw}
To truly understand scenes, 2D representations are insufficient and thus significant strides have been made in recent years to create highly representative 3D representations. That work is mostly centered around static or rigid geometries such as point-clouds \cite{guo2020deeppoint, bello2020deeppoint, wu2019pointconvpoint, tchapmi2017segcloudpoint}, voxels \cite{shin2018voxel, zhou2018voxelnet, wu20153dvoxel, kim20133dvoxel}, octrees \cite{wilhelms1992octrees, yu2021plenoctreesspeed}, or various computed tomography algorithms \cite{buzug2011computedct}. However, these representations are computationally intensive and are seldom extendable to dynamic and non-rigid scenes. Recent advances in implicit functions for representing scenes have shown the ability to implicitly encode photometric attributes such as colour, surface illumination, opacity etc. using shallow neural nets \cite{martin2021nerf, attal2021torfnerf, park2021hypernerf, mildenhall2022nerf}, an important characteristic of these approaches is that they are usually self-supervised. The models can be trained in an end-to-end manner by leveraging a pixel-wise photometric reconstruction loss, which is in strong contrast to existing approaches that require strong priors \cite{deng2022depthprior, roessle2022denseprior, johari2022geonerfprior} or existing templates \cite{park2021nerfies, guo2021template, xu2019deeptemplate, xie2021figtemplates, wei2021nerfingmvstemplate}, each of which are challenging to acquire and significantly limit the generalizability of these approaches to ideal pre-designed scenes. 

Some approaches within the implicit representation paradigm have shown the ability to generate incredibly detailed scenes from a limited set of images \cite{park2021hypernerf, park2021nerfies, li2021neuralnsff, gao2021dynamicnerf}. These immaculate reconstructions illustrate an important proof-of-concept and the expressibility of well-crafted representations, but aren't directly extendable to in-the-wild scenes. This is because the input images, although sparse, are well-posed, capture a $360\deg$ panoramic view of the scene, and {\em must} come with pre-computed pose information. In-the-wild scenes are generally captured from a monocular source and do not include pose information. This information is calculated using off-the-shelf models that use \textit{structure-from-motion} for calculating camera pose and are non-deterministic, and thus prone to error \cite{schonberger2016structureCOLMAP}.

Some approaches extend the implicit representation paradigm to include time $\tau$. These are highly relevant to this paper; specifically, \textit{in-the-wild} approaches that are extremely under-constrained and require  \textit{off-the-shelf} models. Flow-based methods such as \cite{li2021neuralnsff, gao2021dynamicnerf} use off-the-shelf methods to constrain the scene by using additional inputs such as depth estimation \cite{lasinger2019towardsmidas}, optical flow \cite{beauchemin1995optical, teed2020raft}, and semantic segmentation \cite{girshick2018detectron}. Additionally, some work has been done using deformation-based approaches  \cite{pumarola2021dnerf, tretschk2021non} to model dynamic scenes. Although that work is agnostic to input data, and depend on neither pre-trained models nor pre-designed priors, it doesn't generalize well to \textit{in-the-wild} scenes. Our approach combines existing flow-based approaches with deformation-based approaches to set the state-of-the-art on the NVIDIA dynamic scenes dataset.

We also leverage the work done in multi-resolution encoding, which have shown the ability to significantly improve reconstructions by encoding data as a multi-resolution subset of high-frequency embeddings, as measured by commonly-used reconstruction metrics, Learned Perceptual Image Patch Similarity (LPIPS) \cite{zhang2018unreasonablelpips}, Structural Similarity (SSIM) \cite{brunet2011mathematicalssim}, and peak Signal-to-noise ratio (PSNR) \cite{huynh2008scopepsnr}. 


\begin{figure*}
  \centering
   \includegraphics[width=\linewidth]{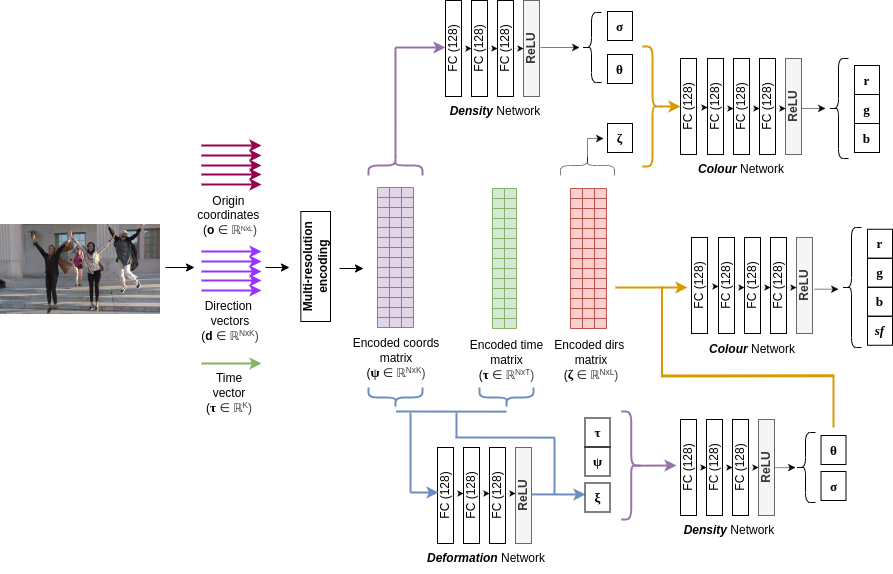}
   \caption{Our proposed end-to-end trainable architecture. We use density and colour networks (\textit{top}) to model a static representation of the scene, and density, colour, and deformation networks (\textit{bottom}) to model the motion-centric pixels in the image. Each set of representations are trained separately and the final image consists of the fused output, as described in section \ref{subsubsec:blending}.}
   \label{fig:arch}
\end{figure*}

\section{Method}
\label{sec:method}
We introduce wildNeRF, a computationally-efficient neural renderer for novel view and time synthesis tasks. The proposed renderer reconstructs complex scenes from sparse inputs captured using a monocular camera source. We build on prior work on flow-based \cite{li2021neuralnsff, gao2021dynamicnerf} and deformation-based \cite{pumarola2021dnerf, tretschk2021non} approaches to model dynamic scenes using computationally efficient grid embeddings \cite{muller2022instantspeed}. These approaches extend traditional neural radiance representations by modelling the change in geometry of objects in the scene over time. This yields high-fidelity reconstructions of complex scenes with varying geometries and intricate dynamics. We illustrate that a combination of deformation and flow-based techniques is able to capture scenes in a holistic manner without loss of generality on novel-view and novel-time synthesis tasks. We leverage \textit{off-the-shelf} pre-trained models for generating flow \cite{teed2020raft} and depth priors \cite{lasinger2019towardsmidas} to constrain the canonical reconstruction space, leading to high-quality reconstructions. We also leverage effective grid encodings \cite{muller2022instantspeed} to produce results in a fraction of the time of existing models.  

\subsection{Datasets}
In this paper, we show quantitative and qualitative results on the NVIDIA Dynamic Scenes Dataset \cite{yoon2020novel}, and  qualitative results (due to the lack of ground truth depth data) on complex surgical datasets, Cholec80 \cite{twinanda2016endonetcholec80} and SugicalActions160 \cite{sa160},  which deal with significant unstructured motion between the camera and the objects in a scene.

\textbf{NVIDIA Dynamic Scenes Dataset}: This dataset consists of 8 scenes, each captured using 12 synchronized cameras to \textit{mimic} motion. In each case, 24 frames from the dataset are held out and used for training and 12 are set aside for testing.

\textbf{Cholec80}: This dataset contains 80 full-length surgical videos capturing the \textit{cholecystectomy} procedure. The scope of existing neural radiance field approaches are limited since they can only model very constrained scenes consisting of a small set of frames. We thus focus on short clips extracted from the longer videos to evaluate the efficacy of these approaches in extremely unconstrained environments.

\textbf{SurgicalActions160}: This dataset consists of a total of 160 short clips ($\sim$5 seconds) in which surgeons perform critical actions such as \textit{suturing}, \textit{dissection}, and \textit{knotting}. 

\subsection{Scene registration}
To completely learn a scene, we require relative camera pose information as input along with image coordinates. Extracting this information from a set of sparse images is called `scene registration', where the structure of features within a scene are used to generate the relative pose characteristics of the constituent images, this is a well studied problem in computer vision and visual perception, called \textit{structure-from-motion}. We use COLMAP \cite{schonberger2016structureCOLMAP}, an open-source structure-from-motion library, for this purpose. 

\subsection{Volume rendering}
We build on the original NeRF \cite{mildenhall2021nerf} formulation and describe the volume rendering equations used in our analysis. We parameterize a ray extending from its point of origin \textbf{o}, with direction \textbf{d} as follows: $y(\textit{b}) = \textbf{o} + \textit{b}.\textbf{d}$. Where \textit{b} refers to a uniformly sampled point along the ray that exists within the near and far bounds, \textbf{$b_{n}$} and \textbf{$b_{f}$} respectively. Using this formulation, we can define a generalized version of the the expected colour \textbf{C} of a pixel as follows:

\begin{equation}
    C(\Theta) = \int_{b_{n}}^{b_{f}} 
                                \textcolor{red}{\underbrace{\tau(b,\Theta)}_\textrm{Transmissivity}}
                                \textcolor{blue}{\underbrace{\sigma(\textbf{p}(b, \Theta))}_\textrm{Pixel density}}
                                \textcolor{brown}{\underbrace{c(\textbf{p}(b, \Theta), d)}_\textrm{Pixel colour}}
                                db
\label{eq:vr}        
\end{equation}
The transmittance $\tau$ defined in eq.\ref{eq:vr} is the probability that a ray between points \textit{$b_{n}$} and \textit{$b$} is absorbed, scattered, or reflected by other objects in its path. 
\begin{equation}
     \textcolor{red}{\tau (b, \Theta)} = e^{\int_{b_{n}}^{b} \sigma(p(b, \Theta)) db}
\label{eq:transmissivity}        
\end{equation}
\begin{equation}
     \textcolor{blue}{\sigma(\textbf{p}(b,\Theta))} = f_{\phi}(\textbf{p}(b,\Theta))
\label{eq:sigma}
\end{equation}
\begin{equation}
     \textcolor{brown}{c(\textbf{p}(b,\Theta), \textbf{d})} = f_{\theta}(\textbf{p}(b,\Theta), \textbf{d})
\label{eq:colour}    
\end{equation}

Equations \ref{eq:sigma} and \ref{eq:colour} are outputs of fully-connected models, details of which are depicted in Figure \ref{fig:arch}. $\Theta$ is the set of variables that parameterize the model. When modelling rigid or non-rigid body motion, $\Theta = (x, d, t)$ where $x$, $d$ and $t$ refer to the input points, input directions, and input time, respectively. As in previous work \cite{mildenhall2021nerf}, we use numerical quadrature \cite{lyness1967numerical} to approximate these integrals. We uniformly sample a set of evenly-spaced points between the near and far bounds, \textbf{$b_{n}$} and \textbf{$b_{f}$}. The resulting equations are as follows:

\begin{equation}
    C(\Theta) = \sum_{b=1}^{N} 
                            \textcolor{red}{\underbrace{\tau (b, \Theta)}_\textrm{Transmissivity}}  
                            \textcolor{blue}{\underbrace{\alpha(\textbf{p}(b, \Theta, \delta))}_\textrm{Pixel density}}
                            \textcolor{brown}{\underbrace{c(\textbf{p}(b, \Theta), d)}_\textrm{Pixel colour}}
\label{eq:vr-nq}        
\end{equation}
\begin{equation}
    \textcolor{red}{\tau (b, \Theta)} = e^{-\sum_{b_{n}}^{b} \sigma(\textbf{p}(b, \Theta)) \delta_{n}}
\label{eq:transmissivity-nq}        
\end{equation}
\begin{equation}
    \textcolor{blue}{\alpha (b, \Theta, \delta)} = 1- e^{-\sigma(\textbf{p}(b, \Theta)) \delta_{n}},
\label{eq:transmissivity-nq-1}        
\end{equation}
where \textcolor{blue}{$\sigma$} and \textcolor{brown}{c} are model outputs (Figure \ref{fig:arch}), and $\delta_{n}=b_{n+1}-b_{n}$ is the distance between two consecutive quadrature points.

\subsection{Pixel pruning}
\label{sec:pruning}
\begin{figure}[ht]
    \centering
    \includegraphics[width=0.5\linewidth]{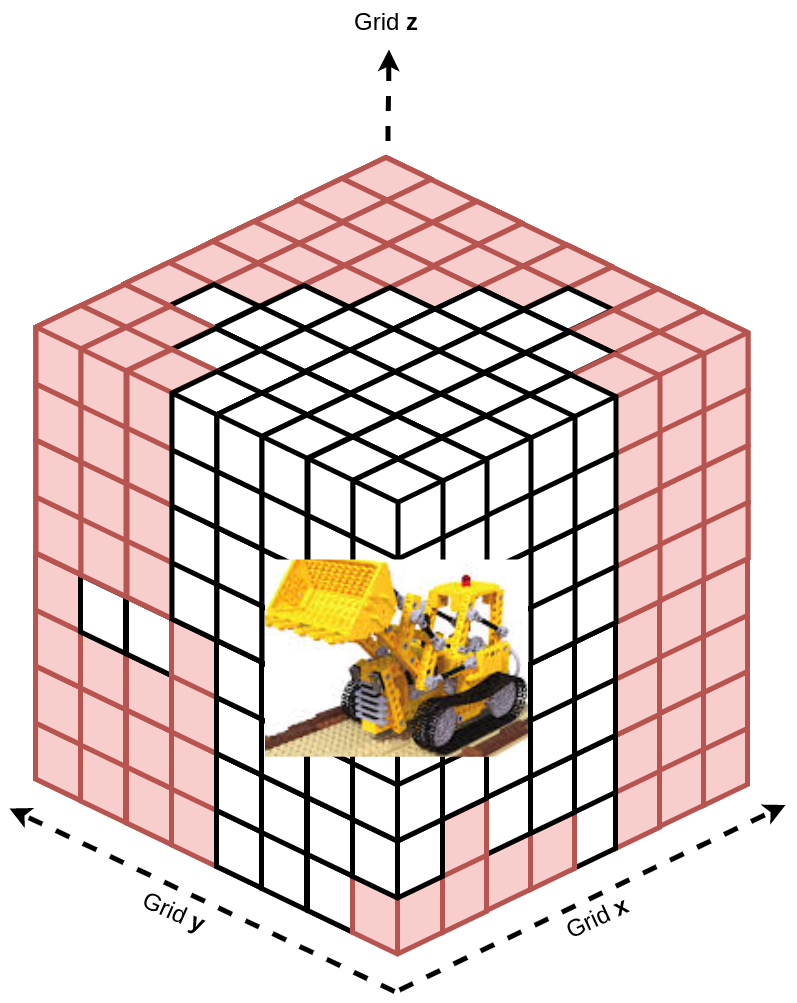}
    \caption{Illustration of the pixel pruning process. Over the first few iterations of the model, any pixel on our grid illustrated as {\color{red}{red}} is excluded from the analysis.}
    \label{fig:grid}
\end{figure}

Traditional approaches \cite{martin2021nerf, mildenhall2021nerf, mildenhall2022nerf} have used  strategically designed coarse and fine sampling schedules of subtended rays for calculating colour pixel values, eq. \ref{eq:vr-nq}. Our approach is inspired by recent advances in efficient training methods of voxel based models \cite{schwarz2022voxgraf, liu2020neural, martel2021acorn} in which we systematically \textit{turn-off} points in space that don't contribute to our scene for a fixed number of iterations. In doing so, we significantly reduce the number of matrix operations required in downstream calculations. After $N=5000$ iterations, we continue optimizing the most \textit{representative} pixels in the image. We illustrate our approach in Figure \ref{fig:grid}

\begin{figure}[h]
    \centering
    \includegraphics[width=1.0\linewidth]{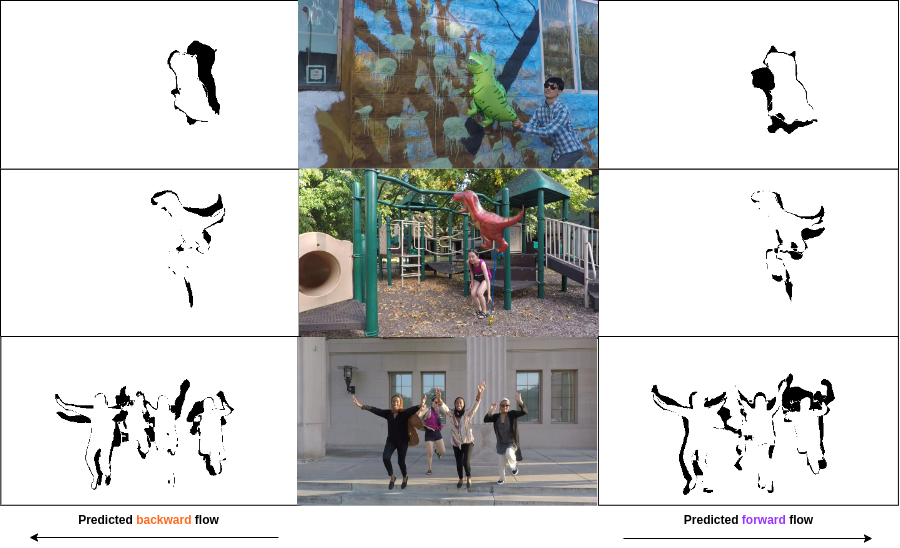}
    \caption{Predicted flow across frames generated by the scene-flow model. These predictions, in conjunction with off-the-shelf estimations of rigid-body flow, help constrain our extremely under-constrained scene.}
    \label{fig:flow}
\end{figure}

\subsection{Model architecture}
Our proposed renderer consists of 3 core lightweight fully-connected models, a \textit{flow model} for predicting the scene-flow, a \textit{deformation model} for modelling non rigid-body dynamics of the model from one frame, to the next, and a \textit{background model} that is agnostic of any motion-centric pixels in the scene. We illustrate the two predictive outputs and how they are used for training the model in an end-to-end manner in Figure \ref{fig:arch}.

\subsubsection{Background Network}
We model the pixels associated with rigid bodies in a scene using a conventional radiance-based implicit representation model \cite{mildenhall2021nerf}. We mask the motion-centric pixels and use numeric quadrature approximations (Eq. \ref{eq:vr-nq}) to map the canonical scene coordinates to the corresponding 2D pixels. For a fair comparison against results in literature, we use $N=256$ uniformly sampled points along each ray. The static paradigm can thus be represented as follows:

\begin{equation}
    \sigma^{static} = f_{\theta}(r(p_{n}))  
\end{equation}
\begin{equation}
    c^{static} = f_{\phi}(r(p_{n}), d),  
\end{equation}
where the functions $f_{\theta}$ and $f_{\phi}$ are represented using the static fully connected models illustrated in Figure \ref{fig:arch}.
\begin{equation}
    C^{s}(x, d) = \sum_{b=1}^{N} 
                            \underbrace{\tau (b, x, d)}_\textrm{Transmissivity}  
                            \underbrace{\alpha(p(b, x, d, \delta))}_\textrm{Pixel density}
                            \underbrace{c(p(b, x, d), d)}_\textrm{Pixel colour}
\label{eq:static}
\end{equation}
\begin{equation}
    \begin{aligned}
    \textcolor{red}{\mathcal{L}^{s}} =
    \sum_{n} || C^{s} - C^{gt} ||_{2}^{2}
    \end{aligned}
\label{eq:static-loss}
\end{equation}
where $M$ is pre-calculated pixel mask that denotes the presence of motion-centric pixels in the training data. 

\subsubsection{Deformation network}
We model non-rigid motion across frames by defining a transformation $\Delta x$, that is applied to points in canonical space across different instances in time. This deformation, $\phi$, jointly with flow-based rigid body motion modelling allows us to improve on existing state-of-the-art results on the NVIDIA dynamic scenes dataset. This transformation is learned by minimizing the mean squared error over the entire dataset.

\begin{equation}
    \Delta x = f_{\psi} (x, d, t)
\end{equation}
\begin{equation}
    x^{*} = x + \Delta x,
\end{equation}
where $x^{*}$ is our predicted non-rigid body deformation, which is included in our downstream analysis. 

\subsubsection{Scene-flow network}
As is the case when attempting to model a large set, of variables using the plenoptic function, the scene becomes increasingly under-constrained. We address these anomalies by leveraging photometric constraints to model rigid-body motion of dynamic pixels in our scene. We model the flow field of the non-rigid motion across adjacent frames in both the forward ($t_{c} = t+1$) and backward ($t_{c} = t-1$) directions (Figure \ref{fig:flow}). The flow field is determined as follows:

\begin{equation}
    (sf_{b}, sf_{f}, \sigma_{t}^{dynamic}, c_{t}^{dynamic}) = f_{\phi} (x^{*}, d, t)
\end{equation}
\begin{equation}
    (sf_{t-1}^{b}, sf_{t-1}^{f}, \sigma_{t-1}^{dynamic}, c_{t-1}^{dynamic}) = f_{\phi} (x^{*}, d, t-1)
\end{equation}
\begin{equation}
    (sf_{t+1}^{b}, sf_{t+1}^{f}, \sigma_{t+1}^{dynamic}, c_{t+1}^{dynamic}) = f_{\phi} (x^{*}, d, t+1),
\end{equation}
where $x^{*}$, $d$, and $t$ refer to the pixels in canonical space, their direction vectors, and instances in time.

We extend eq. \ref{eq:static} to the case of rigid-body motion as follows:
\begin{equation}
    C^{d}(\Theta, t) = \sum_{b=1}^{N} 
                             \underbrace{\tau (b, \Theta, t)}_\textrm{Transmissivity}  
                             \underbrace{\alpha(p(b, \Theta, t, \delta))}_\textrm{Pixel density}
                             \underbrace{c(p(b, \Theta, t), d)}_\textrm{Pixel colour}
\end{equation}
\begin{equation*}
    where, \Theta=(x^{*},d)
\end{equation*}

The subsequent fields predicted by the model are compared against the corresponding ground truth images (\textbf{$C^{gt}$}). The overall scene-flow photometric loss is represented as:

\begin{equation}
    \begin{aligned}
    \textcolor{blue}{\mathcal{L}^{sf+\psi}} = 
        \sum_{n=1}^{N} 
            (|| C_{t}^{d} - C_{t}^{gt} ||_{2}^{2} + 
             || C_{t-1}^{d} - C_{t-1}^{gt} ||_{2}^{2} + \\
             || C_{t+1}^{d} - C_{t+1}^{gt} ||_{2}^{2})
    \end{aligned}
\label{eq:sf-2}
\end{equation}

The scene flow loss is carefully designed to constrain rigid motion across frames eq. \ref{eq:sf}. We illustrate the predicted scene-flow in Figure \ref{fig:flow}.

\begin{figure}[ht]
    \centering
    \includegraphics[width=1.0\linewidth]{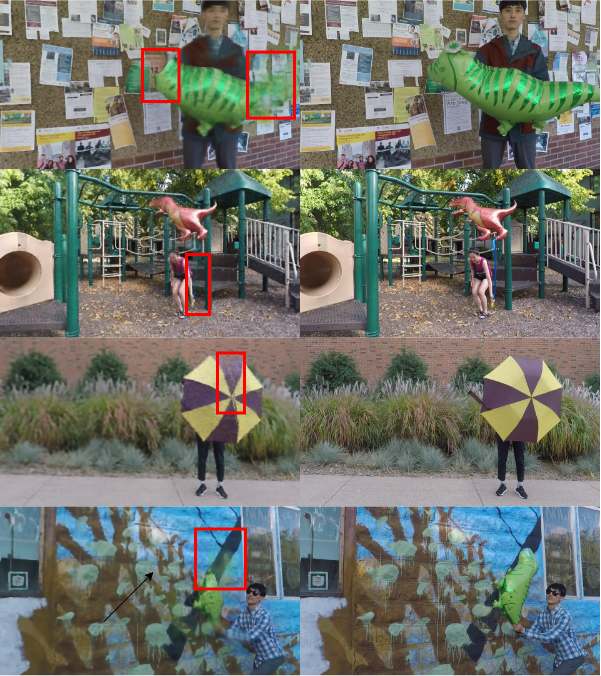}
    \caption{Errors encountered during novel-view synthesis reconstruction.}
    \label{fig:fumbles}
\end{figure}

\begin{figure*}
\centering
\begin{subfigure}{.12\linewidth}
    \includegraphics[width=\linewidth]{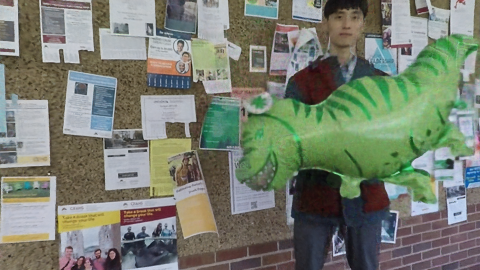}
\end{subfigure}
\begin{subfigure}{.12\linewidth}
    \includegraphics[width=\linewidth]{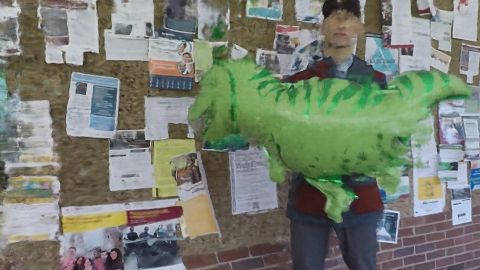}
\end{subfigure}
\begin{subfigure}{.12\linewidth}
    \includegraphics[width=\linewidth]{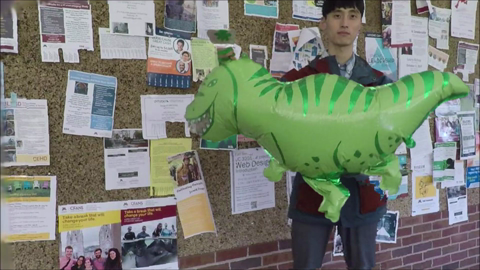}
\end{subfigure}
\begin{subfigure}{.12\linewidth}
    \includegraphics[width=\linewidth]{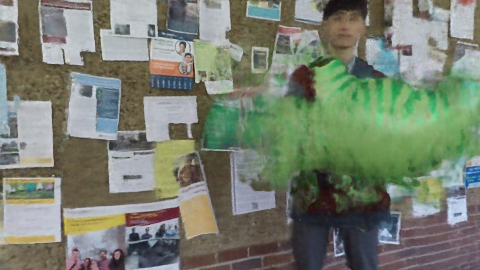}
\end{subfigure}
\begin{subfigure}{.12\linewidth}
    \includegraphics[width=\linewidth]{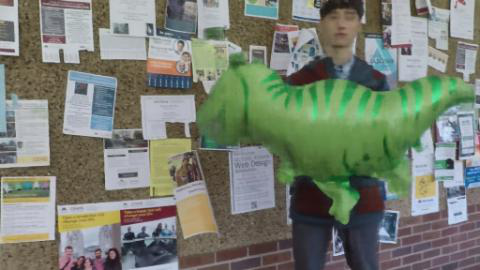}
\end{subfigure}
\begin{subfigure}{.12\linewidth}
    \includegraphics[width=\linewidth]{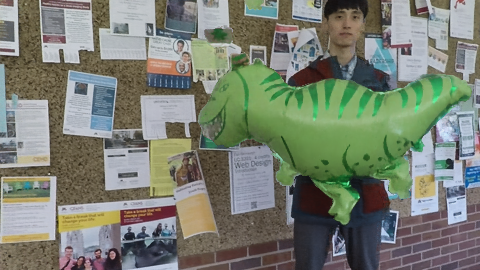}
\end{subfigure}
\begin{subfigure}{.12\linewidth}
    \includegraphics[width=\linewidth]{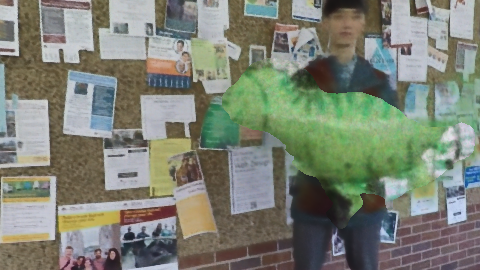}
\end{subfigure}
\begin{subfigure}{.12\linewidth}
    \includegraphics[width=\linewidth]{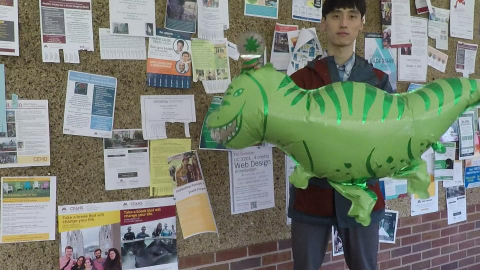}
\end{subfigure}
\begin{subfigure}{.12\linewidth}
    \includegraphics[width=\linewidth]{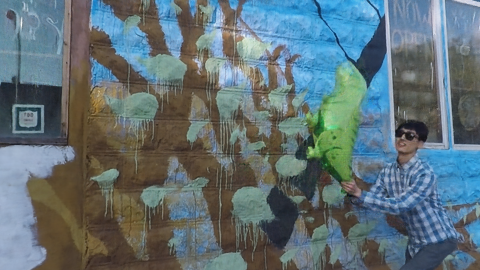}
\end{subfigure}
\begin{subfigure}{.12\linewidth}
    \includegraphics[width=\linewidth]{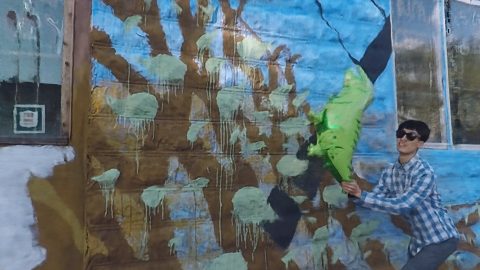}
\end{subfigure}
\begin{subfigure}{.12\linewidth}
    \includegraphics[width=\linewidth]{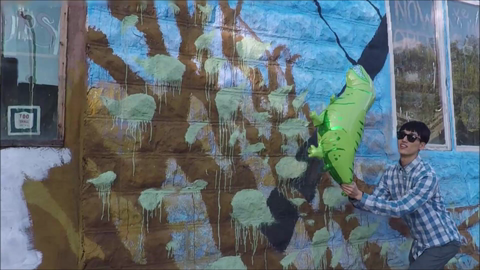}
\end{subfigure}
\begin{subfigure}{.12\linewidth}
    \includegraphics[width=\linewidth]{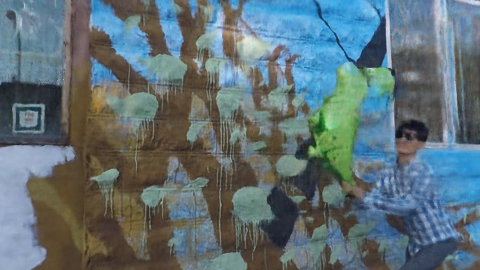}
\end{subfigure}
\begin{subfigure}{.12\linewidth}
    \includegraphics[width=\linewidth]{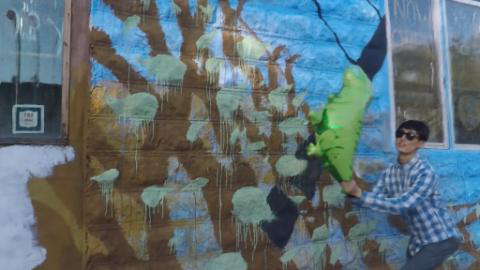}
\end{subfigure}
\begin{subfigure}{.12\linewidth}
    \includegraphics[width=\linewidth]{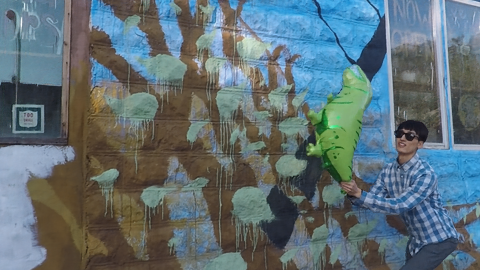}
\end{subfigure}
\begin{subfigure}{.12\linewidth}
    \includegraphics[width=\linewidth]{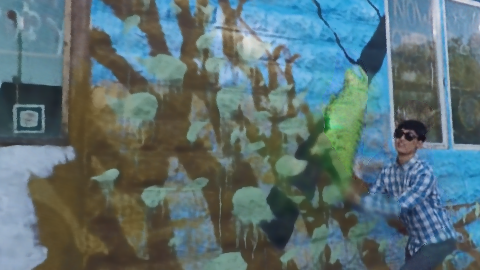}
\end{subfigure}
\begin{subfigure}{.12\linewidth}
    \includegraphics[width=\linewidth]{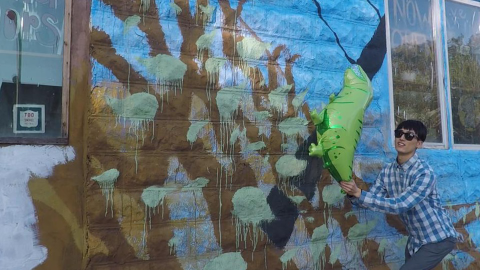}
\end{subfigure}

\begin{subfigure}{.12\linewidth}
    \includegraphics[width=\linewidth]{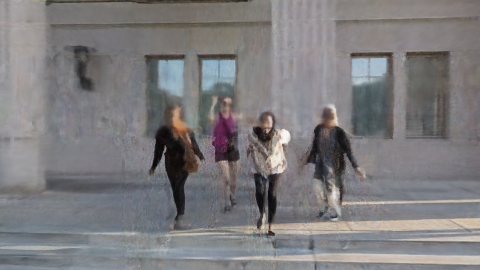}
\end{subfigure}
\begin{subfigure}{.12\linewidth}
    \includegraphics[width=\linewidth]{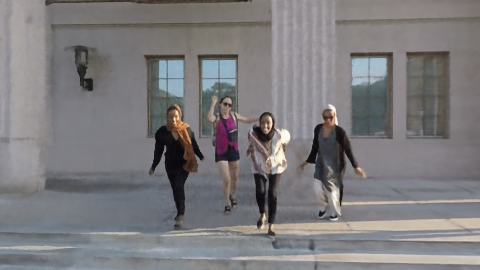}
\end{subfigure}
\begin{subfigure}{.12\linewidth}
    \includegraphics[width=\linewidth]{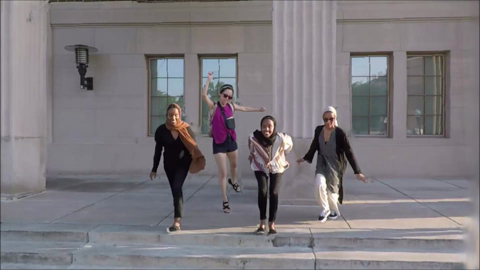}
\end{subfigure}
\begin{subfigure}{.12\linewidth}
    \includegraphics[width=\linewidth]{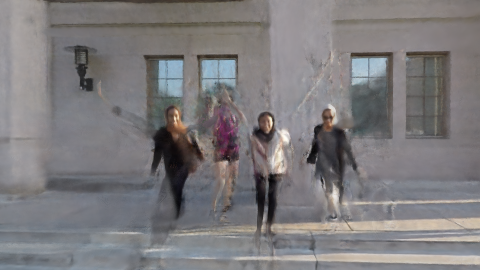}
\end{subfigure}
\begin{subfigure}{.12\linewidth}
    \includegraphics[width=\linewidth]{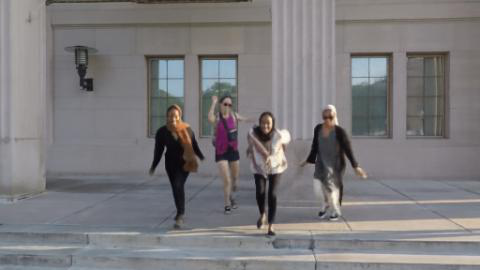}
\end{subfigure}
\begin{subfigure}{.12\linewidth}
    \includegraphics[width=\linewidth]{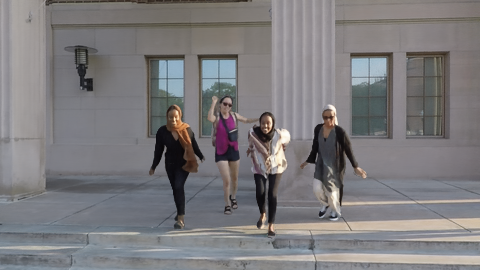}
\end{subfigure}
\begin{subfigure}{.12\linewidth}
    \includegraphics[width=\linewidth]{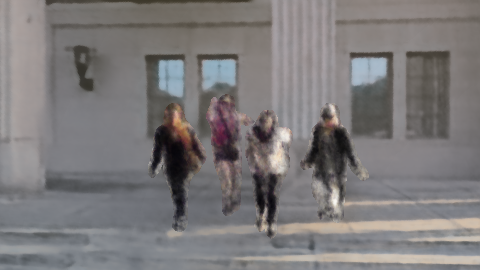}
\end{subfigure}
\begin{subfigure}{.12\linewidth}
    \includegraphics[width=\linewidth]{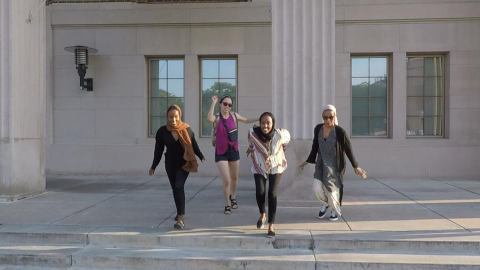}
\end{subfigure}

\begin{subfigure}{.12\linewidth}
    \includegraphics[width=\linewidth]{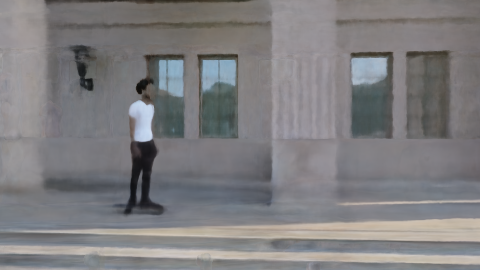}
\end{subfigure}
\begin{subfigure}{.12\linewidth}
    \includegraphics[width=\linewidth]{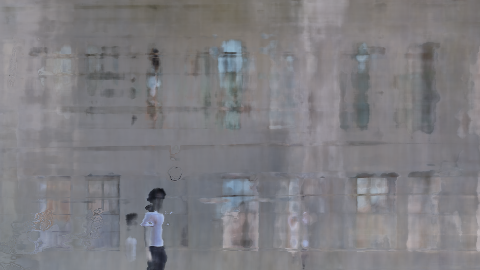}
\end{subfigure}
\begin{subfigure}{.12\linewidth}
    \includegraphics[width=\linewidth]{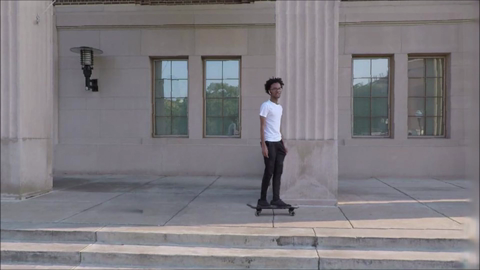}
\end{subfigure}
\begin{subfigure}{.12\linewidth}
    \includegraphics[width=\linewidth]{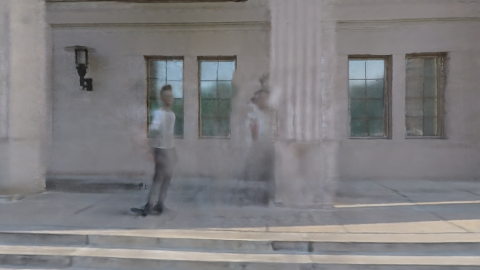}
\end{subfigure}
\begin{subfigure}{.12\linewidth}
    \includegraphics[width=\linewidth]{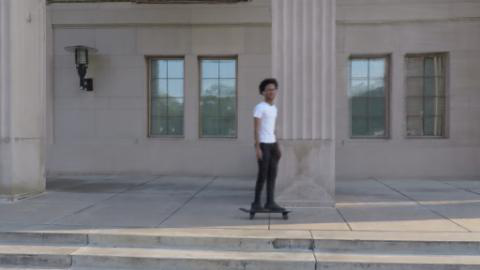}
\end{subfigure}
\begin{subfigure}{.12\linewidth}
    \includegraphics[width=\linewidth]{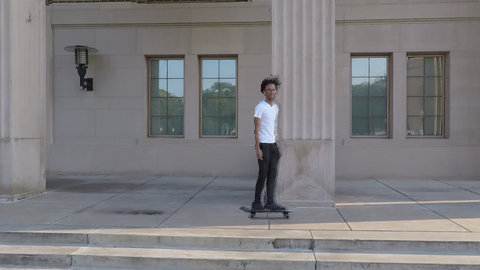}
\end{subfigure}
\begin{subfigure}{.12\linewidth}
    \includegraphics[width=\linewidth]{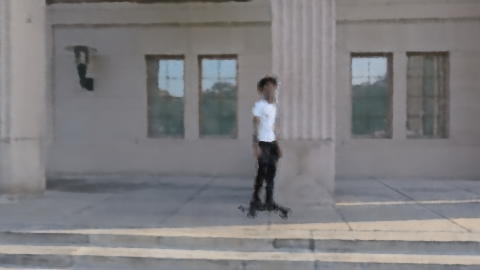}
\end{subfigure}
\begin{subfigure}{.12\linewidth}
    \includegraphics[width=\linewidth]{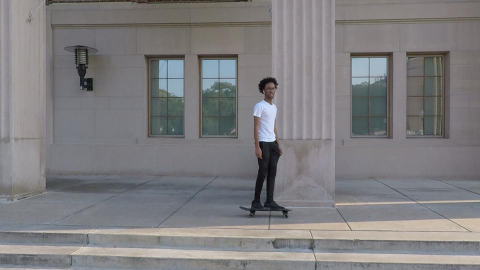}
\end{subfigure}

\begin{subfigure}{.12\linewidth}
    \includegraphics[width=\linewidth]{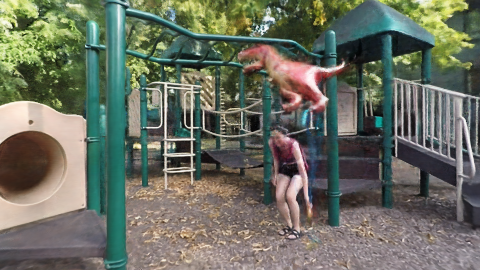}
\end{subfigure}
\begin{subfigure}{.12\linewidth}
    \includegraphics[width=\linewidth]{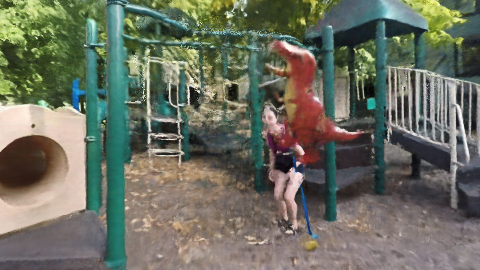}
\end{subfigure}
\begin{subfigure}{.12\linewidth}
    \includegraphics[width=\linewidth]{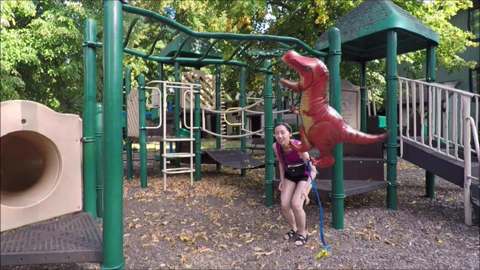}
\end{subfigure}
\begin{subfigure}{.12\linewidth}
    \includegraphics[width=\linewidth]{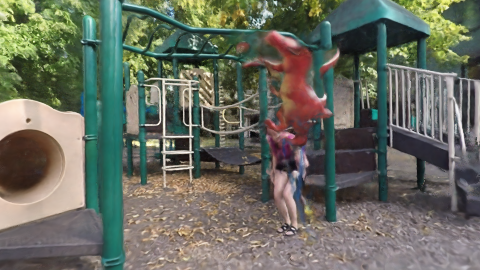}
\end{subfigure}
\begin{subfigure}{.12\linewidth}
    \includegraphics[width=\linewidth]{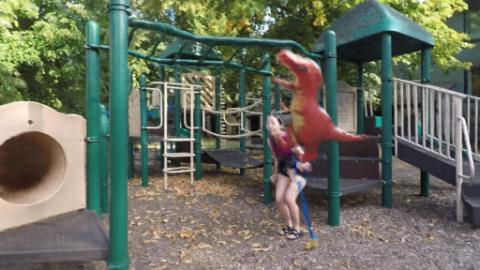}
\end{subfigure}
\begin{subfigure}{.12\linewidth}
    \includegraphics[width=\linewidth]{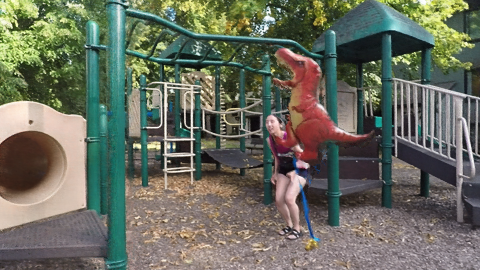}
\end{subfigure}
\begin{subfigure}{.12\linewidth}
    \includegraphics[width=\linewidth]{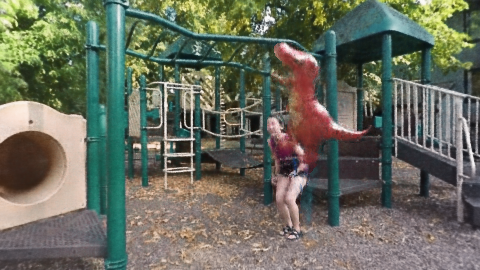}
\end{subfigure}
\begin{subfigure}{.12\linewidth}
    \includegraphics[width=\linewidth]{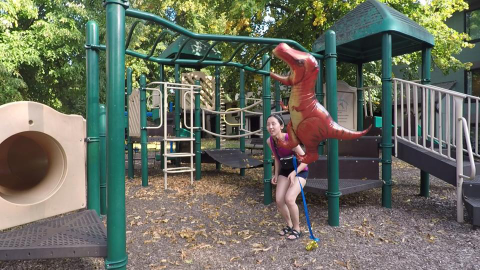}
\end{subfigure}

\begin{subfigure}{.12\linewidth}
    \includegraphics[\empty width=\linewidth]{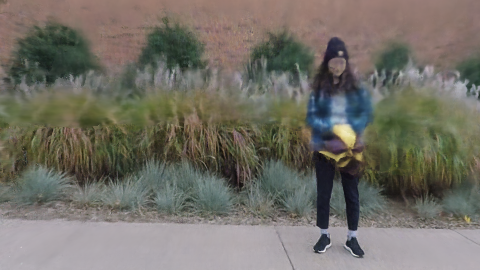}
    \caption{\textit{NeRF}}
\end{subfigure}
\begin{subfigure}{.12\linewidth}
    \includegraphics[\empty width=\linewidth]{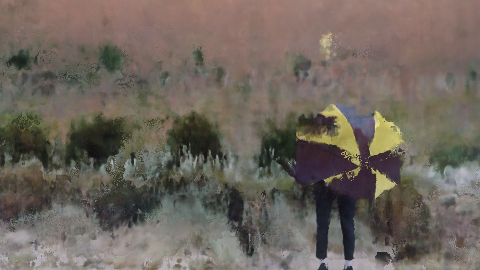}
    \caption{\textit{NeRF+time}}
\end{subfigure}
\begin{subfigure}{.12\linewidth}
    \includegraphics[\empty width=\linewidth]{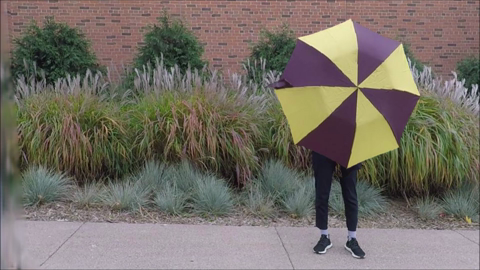}
    \caption{\textit{Yoon et al.}}
\end{subfigure}
\begin{subfigure}{.12\linewidth}
    \includegraphics[\empty width=\linewidth]{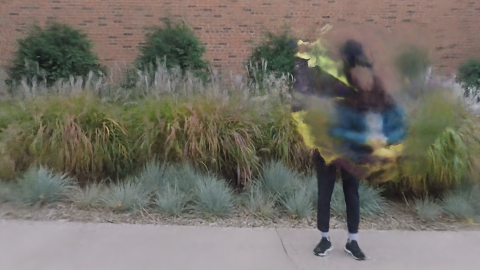}
    \caption{\textit{Tretschk et al.}}
\end{subfigure}
\begin{subfigure}{.12\linewidth}
    \includegraphics[\empty width=\linewidth]{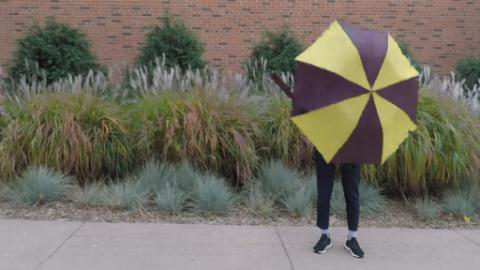}
    \caption{\textit{Li et al.}}
\end{subfigure}
\begin{subfigure}{.12\linewidth}
    \includegraphics[\empty width=\linewidth]{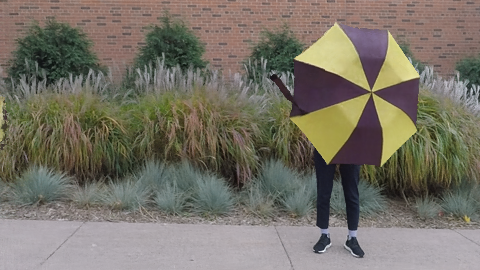}
    \caption{\textit{Gao et al.}}
\end{subfigure}
\begin{subfigure}{.12\linewidth}
    \includegraphics[\empty width=\linewidth]{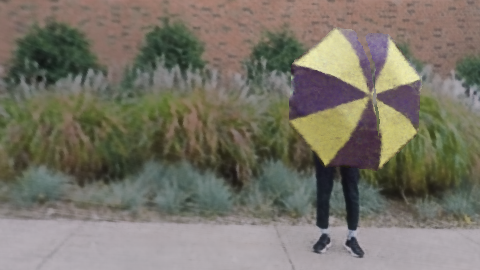}
    \caption{Ours}
\end{subfigure}
\begin{subfigure}{.12\linewidth}
    \includegraphics[\empty width=\linewidth]{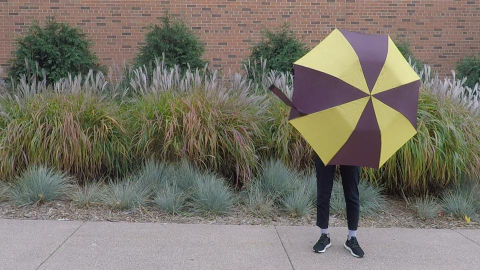}
    \caption{Ground Truth}
\end{subfigure}

\caption[short]{\textbf{Qualitative results}: We compare our qualitative results to existing approaches and notice that our results are competitive with existing \textit{state-of-the-art} approaches such as \textit{NSFF} and \textit{Dynamic NeRF} despite only being trained for a fraction of the time. We compare and contrast the training speeds in Table \ref{tab:train-speed}}
\label{fig:qualitative}
\end{figure*}

\begin{table}
  \centering
  \begin{tabular}{@{}lcc@{}}
    \toprule
    Method & Train Time (hours) $\downarrow$ & Speedup $\uparrow$ \\
    \midrule
    NeRF \cite{mildenhall2021nerf} (\textit{naive}) & ~48 (0.1) & \color{green}{x32} \\
    Tretschk et al. \cite{tretschk2021non} &  ~15 (0.1) & \color{green}{5x}\\
    Li et al. \cite{li2021neuralnsff} & ~50 (0.3) & \color{green}{x35}\\
    Gao et al. \cite{gao2021dynamicnerf} & 48 (0.3) & \color{green}{x33}\\
    Ours (\textit{w/o pruning}) & 1.7 (0.1)  & \color{green}{x1.1}\\
    Ours & 1.6 (0.1) & - \\
    \bottomrule
  \end{tabular}
  \caption{All experiments were done using an RTX3090 GPU. We present the processing times required to produce the reported PSNR values in Table \ref{tab:main}. The speedup column indicates the speedup of the proposed method compared to \textit{our} method, both with and without pixel pruning \ref{sec:pruning}.}
  \label{tab:train-speed}
\end{table}

\subsubsection{Blending approach}
\label{subsubsec:blending}
Our blending approach utilizes the readily available motion masks generated during the pre-processing step and assigns pixel-level values depending on the masks value. This blending approach is simple and doesn't require additional networks \cite{pumarola2021dnerf} to model the background pixels. It also avoids a \textit{blending} parameter that is computed in addition to other network outputs as in \cite{li2021neuralnsff, gao2021dynamicnerf}, each of which require significantly higher computational times and trainable parameters. The final image is computed as follows:

\[
    C(p)= 
    \begin{cases}
        C(p)^{dynamic},& \text{if } M\geq 0.5\\
        C(p)^{static},              & \text{otherwise}
    \end{cases}
\]

\subsection{Training Details}
\label{subsec:training}

For training, we use the pixel-level motion masks to delineate between the static and motion-centric pixels (caused by both rigid and non-rigid motion) in the image. Each image was resized to $480\times 270$ to remain consistent with previous approaches. We sample 
a set of $N=4096$ rays subtending from these pixels and use them to train to separate models, a simple static or background model, inspired by the original NeRF \cite{mildenhall2021nerf} formulation, and a dynamic or motion-centric model, consisting of an in-built deformation network. This schema allowed us to run all our experiments on an RTX3090 GPU with 24Gb of VRAM. We use an ADAM optimized training scheme, with $l_{r}^{network}=1e-3$ and $l_{r}^{model}=1e-2$. We also use a learning rate scheduler and an exponential decay value of $5e-5$. Finally, our overall loss is calculated as follows:
\begin{equation}
    \begin{aligned}
    \mathcal{L}^{total} = \textcolor{red}{\mathcal{L}^{s}} + \textcolor{blue}{\mathcal{L}^{sf+\psi}}
    \end{aligned}
\label{eq:sf}
\end{equation}


\section{Results} 
\subsection{Experimental setup}
We evaluate our proposed algorithm on the Dynamic Scene Dataset \cite{yoon2020novel}. The dataset consists of 9 video sequences each captured simultaneously using 12 fixed cameras at 12 time steps. The input for each model is a set of frames ${I_{ct}}$, where \textbf{c} refers to the camera pose and \textbf{t} refers to an instance in time. Figure \ref{fig:arch} illustrates the problem setup as we take individual frames at different points in time and attempt to create a generalizable model for predicting novel-view and novel-time states, should one of the controllable parameters $(c,t)$ be fixed. We use COLMAP \cite{schonberger2016structureCOLMAP}, a commonly used \textit{structure-to-motion} model that leverages structural information captured from disjointed frames to create a relative mapping in an unsupervised manner, thus yielding pose information for our input frames. Our setup also uses motion-masking to differentiate between static and motion-centric pixels similar to Gao \textit{et al} \cite{gao2021dynamicnerf}. We discuss our quantitative and qualitative results in the sections below.

\subsection{Quantitative}
We present our quantitative results against state-of-the-art benchmarks in Table \ref{tab:main}. We observe improvements of $~3\%$ and $~1\%$ in Peak signal-to-noise ratio (PSNR - $\uparrow$) and Structural Similarity (SSIM - $\uparrow$) respectively, despite only training for a fraction of the time as earlier works \cite{li2021neuralnsff, gao2021dynamicnerf, mildenhall2021nerf}. Additionally, we present a comparison of the training times required by existing approaches in Table \ref{tab:train-speed} and show that \textit{pixel pruning} (Section \ref{sec:pruning}) and a grid generated using a combination of frequency encoding \cite{mildenhall2021nerf} and multi-resolution hash encoding \cite{muller2022instantspeed} generates results in unparalleled times on complex dynamic scenes. We also observe that most scenes samples are competitive against existing approaches except the \textit{Truck} scene. We believe that this is due to inconsistent camera pose parameters generated by COLMAP. Dissimilar to other scenes, both the \textit{Truck} and \textit{DynamicFaces} scenes required multiple runs of COLMAP as it would regularly fail, as has been documented in earlier works \cite{li2021neuralnsff, gao2021dynamicnerf}. This inconsistent behaviour is an unfortunate by-product of the structure-from-motion paradigm and is an active area of research \cite{lin2021barf, martin2021nerf, wang2021nerf}.         

\subsection{Qualitative}
We present our qualitative results against state-of-the-art benchmarks in Figure \ref{fig:qualitative}. Our results are highly competitive against earlier approaches and avoid significant artifacts due to motion, such as the entire object disappearing, or even complete reconstruction failure. Gao \textit{et al.} \cite{gao2021dynamicnerf} have shown some of the most impressive results on this dataset, yet they note in the paper that their approach leads to artifacts when modelling non-rigid motion. This can be seen in (Figure \ref{fig:qualitative} - column \textit{$(f)$}). Our approach is not only able to model both rigid and non-rigid motion but is also able to capture light-based effects well (Figure \ref{fig:qualitative} - column \textit{$(g)$}).  

\begin{table*}[]
\begin{center}
\end{center}
\scalebox{0.7}{
\begin{tabular}{ccccccccccccc}
\toprule
\multicolumn{1}{l}{} & \multicolumn{3}{c}{\textbf{Jumping}}                                                                                  & \multicolumn{3}{c}{\textbf{Skating}}                                                                                  & \multicolumn{3}{c}{\textbf{Truck}}                                                                                    & \multicolumn{3}{c}{\textbf{Umbrella}}                                                                                 \\
\toprule
\multicolumn{1}{l}{} & PSNR $(\uparrow)$                                 & SSIM $(\uparrow)$                                  & LPIPS $(\downarrow)$                                 & PSNR $(\uparrow)$                                  & SSIM $(\uparrow)$                                  & LPIPS $(\downarrow)$                                 & PSNR $(\uparrow)$                                  & SSIM $(\uparrow)$                                  & LPIPS $(\downarrow)$                                 & PSNR $(\uparrow)$                                  & SSIM $(\uparrow)$                                  & LPIPS $(\downarrow)$                                 \\
\midrule
NeRF                 & 20.57                                 & 0.681                                 & 0.305                                 & 23.04                                 & 0.777                                 & 0.317                                 & 20.35                                 & 0.673                                 & 0.203                                 & 21.08                                 & 0.543                                 & 0.442                                 \\
NeRF + time          & 16.72                                 & 0.428                                 & 0.489                                 & 19.23                                 & 0.471                                 & 0.543                                 & 15.46                                 & 0.364                                 & 0.363                                 & 17.17                                 & 0.242                                 & 0.752                                 \\
Yoon et al.          & 20.15                                 & 0.616                                 & {\color[HTML]{B7B7B7} \textbf{0.148}} & 21.75                                 & 0.561                                 & {\color[HTML]{B45F06} \textbf{0.135}} & 21.53                                 & 0.626                                 & {\color[HTML]{B7B7B7} \textbf{0.099}} & 20.35                                 & 0.543                                 & {\color[HTML]{B7B7B7} \textbf{0.179}} \\
Tretschk et al.      & 19.37                                 & 0.617                                 & 0.295                                 & 23.28                                 & 0.722                                 & 0.235                                 & 17.11                                 & 0.406                                 & 0.408                                 & 19.26                                 & 0.360                                 & 0.428                                 \\
NSFF (Li et al)      & {\color[HTML]{B7B7B7} \textbf{24.10}} & 0.807                                 & {\color[HTML]{B45F06} \textbf{0.156}} & {\color[HTML]{B45F06} \textbf{28.88}} & {\color[HTML]{B45F06} \textbf{0.884}} & {\color[HTML]{B7B7B7} \textbf{0.135}} & {\color[HTML]{B7B7B7} \textbf{23.33}} & {\color[HTML]{B45F06} \textbf{0.697}} & {\color[HTML]{B45F06} \textbf{0.154}} & {\color[HTML]{B45F06} \textbf{22.56}} & {\color[HTML]{B45F06} \textbf{0.651}} & {\color[HTML]{B45F06} \textbf{0.302}} \\
Dynamic NeRF         & {\color[HTML]{B45F06} \textbf{23.57}} & {\color[HTML]{F1C232} \textbf{0.832}} & {\color[HTML]{F1C232} \textbf{0.096}} & {\color[HTML]{B7B7B7} \textbf{31.92}} & {\color[HTML]{B7B7B7} \textbf{0.948}} & {\color[HTML]{F1C232} \textbf{0.037}} & {\color[HTML]{F1C232} \textbf{25.50}} & {\color[HTML]{F1C232} \textbf{0.785}} & {\color[HTML]{F1C232} \textbf{0.074}} & {\color[HTML]{B7B7B7} \textbf{22.68}} & {\color[HTML]{B7B7B7} \textbf{0.711}} & {\color[HTML]{F1C232} \textbf{0.144}} \\
Ours                 & {\color[HTML]{F1C232} \textbf{26.03}} & {\color[HTML]{B7B7B7} \textbf{0.821}} & 0.287                                 & {\color[HTML]{F1C232} \textbf{32.02}} & {\color[HTML]{F1C232} \textbf{0.957}} & 0.206                                 & {\color[HTML]{B45F06} \textbf{21.89}} & {\color[HTML]{B7B7B7} \textbf{0.730}} & 0.331                                 & {\color[HTML]{F1C232} \textbf{24.99}} & {\color[HTML]{F1C232} \textbf{0.715}} & 0.326                                 \\
\midrule
\multicolumn{1}{l}{} & \multicolumn{3}{c}{\textbf{Balloon1}}                                                                                 & \multicolumn{3}{c}{\textbf{Balloon2}}                                                                                 & \multicolumn{3}{c}{\textbf{Playground}}                                                                               & \multicolumn{3}{c}{\textbf{Overall}}                                                                                  \\
\midrule
\multicolumn{1}{l}{} & PSNR $(\uparrow)$                                 & SSIM $(\uparrow)$                                  & LPIPS $(\downarrow)$                                 & PSNR $(\uparrow)$                                  & SSIM $(\uparrow)$                                  & LPIPS $(\downarrow)$                                 & PSNR $(\uparrow)$                                  & SSIM $(\uparrow)$                                  & LPIPS $(\downarrow)$                                 & PSNR $(\uparrow)$                                  & SSIM $(\uparrow)$                                  & LPIPS $(\downarrow)$                                 \\
\midrule
NeRF                 & 19.06                                 & {\color[HTML]{B45F06} \textbf{0.686}} & 0.215                                 & {\color[HTML]{B45F06} \textbf{24.06}} & {\color[HTML]{B45F06} \textbf{0.812}} & {\color[HTML]{B7B7B7} \textbf{0.098}} & 20.84                                 & {\color[HTML]{B45F06} \textbf{0.764}} & {\color[HTML]{B45F06} \textbf{0.166}} & 21.28                                 & 0.705                                 & 0.249                                 \\
NeRF + time          & 17.32                                 & 0.387                                 & 0.304                                 & 19.66                                 & 0.531                                 & 0.237                                 & 13.79                                 & 0.179                                 & 0.445                                 & 17.05                                 & 0.372                                 & 0.448                                 \\
Yoon et al.          & 18.74                                 & 0.606                                 & {\color[HTML]{B7B7B7} \textbf{0.179}} & 19.88                                 & 0.416                                 & {\color[HTML]{B45F06} \textbf{0.139}} & 15.08                                 & 0.249                                 & 0.184                                 & 19.64                                 & 0.517                                 & {\color[HTML]{B7B7B7} \textbf{0.152}} \\
Tretschk et al.      & 16.97                                 & 0.338                                 & 0.353                                 & 22.21                                 & 0.706                                 & 0.213                                 & 14.23                                 & 0.185                                 & 0.338                                 & 18.92                                 & 0.476                                 & 0.324                                 \\
NSFF (Li et al)      & {\color[HTML]{B45F06} \textbf{21.35}} & 0.686                                 & 0.224                                 & 24.02                                 & 0.734                                 & 0.229                                 & {\color[HTML]{B45F06} \textbf{20.85}} & 0.706                                 & 0.219                                 & {\color[HTML]{B45F06} \textbf{23.58}} & {\color[HTML]{B45F06} \textbf{0.738}} & {\color[HTML]{B45F06} \textbf{0.203}} \\
Dynamic NeRF         & {\color[HTML]{B7B7B7} \textbf{21.43}} & {\color[HTML]{B7B7B7} \textbf{0.755}} & {\color[HTML]{F1C232} \textbf{0.115}} & {\color[HTML]{B7B7B7} \textbf{26.59}} & {\color[HTML]{B7B7B7} \textbf{0.857}} & {\color[HTML]{F1C232} \textbf{0.052}} & {\color[HTML]{B7B7B7} \textbf{23.74}} & {\color[HTML]{B7B7B7} \textbf{0.852}} & {\color[HTML]{F1C232} \textbf{0.084}} & {\color[HTML]{B7B7B7} \textbf{25.06}} & {\color[HTML]{B7B7B7} \textbf{0.820}} & {\color[HTML]{F1C232} \textbf{0.086}} \\
Ours                 & {\color[HTML]{F1C232} \textbf{24.60}} & {\color[HTML]{F1C232} \textbf{0.782}} & {\color[HTML]{B45F06} \textbf{0.199}} & {\color[HTML]{F1C232} \textbf{26.60}} & {\color[HTML]{F1C232} \textbf{0.859}} & 0.305                                 & {\color[HTML]{F1C232} \textbf{25.46}} & {\color[HTML]{F1C232} \textbf{0.914}} & {\color[HTML]{B7B7B7} \textbf{0.118}} & {\color[HTML]{F1C232} \textbf{25.94}} & {\color[HTML]{F1C232} \textbf{0.825}} & 0.253        \\       \bottomrule
\end{tabular}}
\caption{\textbf{Quantitative results}: We perform extensive quantitative evaluation and show that our model achieves state-of-the-art PSNR $(\uparrow)$ and SSIM $(\uparrow)$ values compared to existing approaches. For each category, results are ranked and colour coded as follows: \color[HTML]{F1C232}{$1^{st}$}, \color[HTML]{B7B7B7}{$2^{nd}$}, \color[HTML]{B45F06}{$3^{rd}$}.}
\label{tab:main}
\end{table*}

\begin{table}
  \centering
  \begin{tabular}{@{}lccc@{}}
    \toprule
    Method & PSNR $\uparrow$ & SSIM $\uparrow$ & LPIPS $\downarrow$ \\
    \midrule
    w/o Background & 11.741 & 0.186 & 0.889 \\
    w/o Flow & 22.827 & 0.761 & 0.205\\
    w/o Deformation & 23.266 & 0.834 & 0.129\\
    All & 25.460 & 0.854 & 0.118\\
    \bottomrule
  \end{tabular}
  \caption{\textbf{Ablation study}: We examine the effect of using our blended approach of modelling each of the \textit{background}, \textit{rigid motion} and \textit{non-rigid motion} separately. These experiments are done on the \textit{Playground} scene in the NVIDIA dynamic scenes dataset.}
  \label{tab:ablation-study}
\end{table}

\subsection{Ablation study}
We perform an ablation study to examine the effect of each component of our proposed model. The results are presented in Table \ref{tab:ablation-study}. The background, flow, and deformation models each play a critical role in modelling the overall motion in these dynamic frames, especially the deformation network. 

\subsection{Limitations}
Although we are able to generate realistic outputs in both the novel-view synthesis and the novel-time paradigms, and provide state-of-the-art results, our proposed approach is prone to artifacts due to the extremely unconstrained nature of the tasks. We present some of the failure cases in Figure \ref{fig:fumbles}.  We address each of the issues from \textit{top} to \textit{bottom}, with the ground truth images being displayed on the right-hand side. \textbf{Balloon1}: The artifact on the \textit{left} is due to improper motion-mask generation, which results in a portion of the dinosaurs head to be cut-off. The artifacts on the right are due to grid errors, which is potentially a by-product of the pixel-pruning step described in section \ref{sec:pruning}. \textbf{Playground}: The woman in the image is holding a ribbon extending from the balloon; this ribbon is only partially visible in our reconstruction, we believe that this is caused by the deformation network, which isn't able to capture the fine detail. The reader may note that other fine objects in the background are represented quite clearly, which indicates that this is most likely caused by incorrect motion-centric modelling. \textbf{Umbrella}: The noticeable streak on the top of the umbrella is caused by pixel-level masking artifacts. \textbf{Balloon2}: The artifact is caused by improper modelling of rigid-body motion at the apex of the balloon's motion. Most of the identified artifacts are caused by improper blending of motion-centric and background pixels which is an interesting avenue for future research. Although the models are competitive in PSNR values against state-of-the-art models, there remains room for further engineering optimizations to further improve reconstruction rates and overall quality in the future.    

\section{Conclusion}
We present a novel neural radiance-based approach for synthesizing scenes from sparse inputs by modelling both rigid body and non rigid-body motion independently. This approach, coupled with a multi-resolution hash encoding grid, yield state-of-the-art results on the NVIDIA dynamic scenes dataset, while being much faster than existing methods.  
We also move away from idealized environments and datasets and show the efficacy of our approach on complex real-world scenes, such as the SurgicalActions160 and Cholec80 datasets. We hope that our work leads to widespread adoption of neural radiance-based techniques, and will make our source code publicly available.

{\small
\bibliographystyle{ieee_fullname}
\bibliography{egbib}
}

\newpage
\section{Supplementary}

\subsection{Implementation}
The code is built on existing repositories and we would like to acknowledge the wonderful contributions made by \cite{queianchen_nerf} and \cite{TensoRF} for their clean implementation of a fast and modular version of the original NeRF repository \cite{mildenhall2021nerf}. We'd also like to acknowledge the contributions of Muller \textit{et al} \cite{tiny-cuda-nn, muller2022instantspeed} for creating fast CUDA powered kernels and renderers for obtaining excellent speeds during training and inference.

\subsection{SurgicalActions160}
\begin{figure*}
\centering


\begin{subfigure}{.33\linewidth}
    \includegraphics[\empty width=\linewidth]{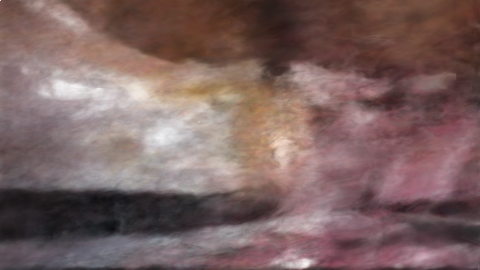}
\end{subfigure}
\begin{subfigure}{.33\linewidth}
    \includegraphics[\empty width=\linewidth]{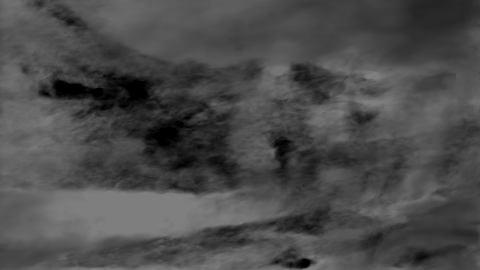}
\end{subfigure}
\begin{subfigure}{.33\linewidth}
    \includegraphics[\empty width=\linewidth]{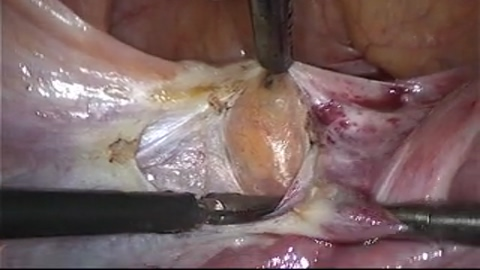}
\end{subfigure}


\begin{subfigure}{.33\linewidth}
    \includegraphics[\empty width=\linewidth]{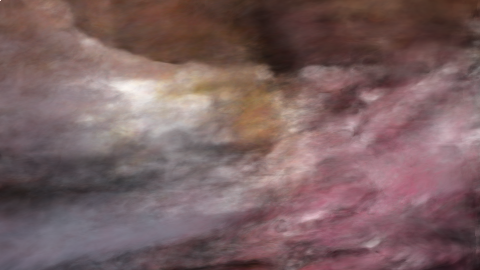}
\end{subfigure}
\begin{subfigure}{.33\linewidth}
    \includegraphics[\empty width=\linewidth]{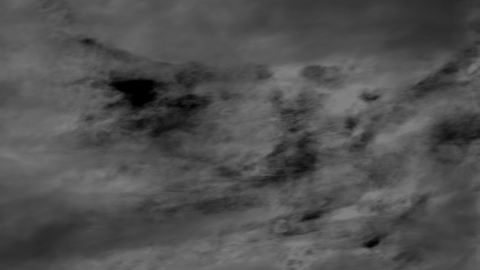}
\end{subfigure}
\begin{subfigure}{.33\linewidth}
    \includegraphics[\empty width=\linewidth]{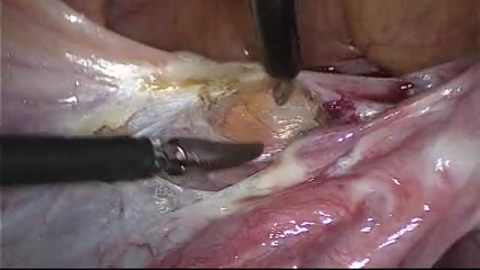}
\end{subfigure}


\begin{subfigure}{.33\linewidth}
    \includegraphics[\empty width=\linewidth]{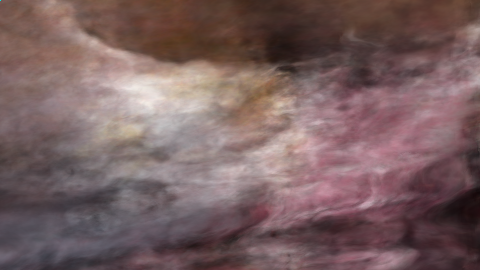}
\end{subfigure}
\begin{subfigure}{.33\linewidth}
    \includegraphics[\empty width=\linewidth]{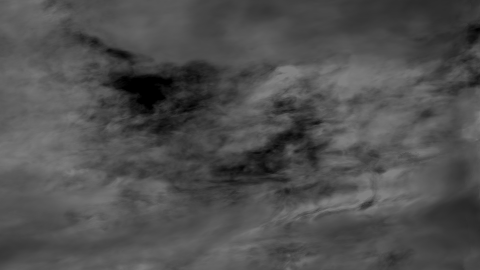}
\end{subfigure}
\begin{subfigure}{.33\linewidth}
    \includegraphics[\empty width=\linewidth]{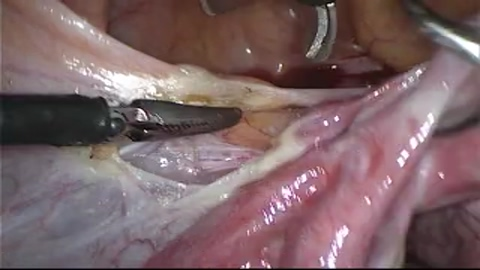}
\end{subfigure}


\begin{subfigure}{.33\linewidth}
    \includegraphics[\empty width=\linewidth]{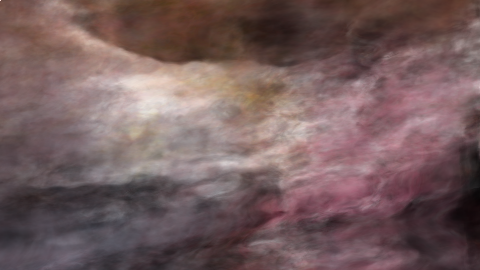}
\end{subfigure}
\begin{subfigure}{.33\linewidth}
    \includegraphics[\empty width=\linewidth]{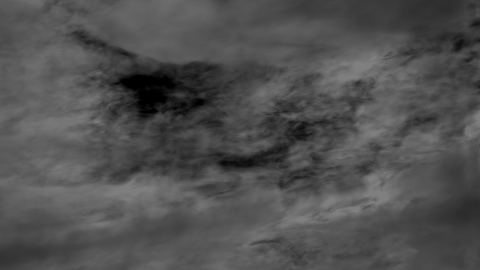}
\end{subfigure}
\begin{subfigure}{.33\linewidth}
    \includegraphics[\empty width=\linewidth]{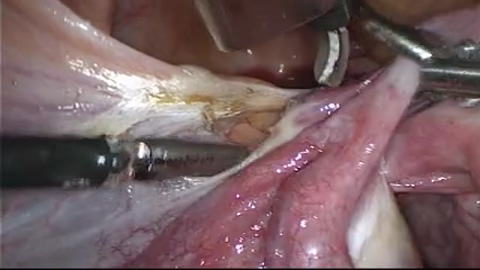}
\end{subfigure}


\begin{subfigure}{.33\linewidth}
    \includegraphics[\empty width=\linewidth]{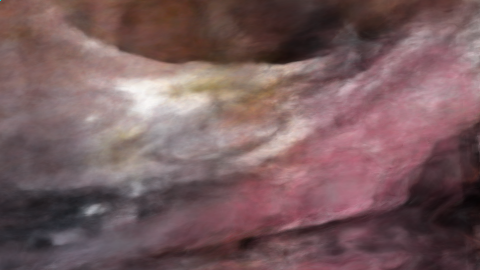}
\end{subfigure}
\begin{subfigure}{.33\linewidth}
    \includegraphics[\empty width=\linewidth]{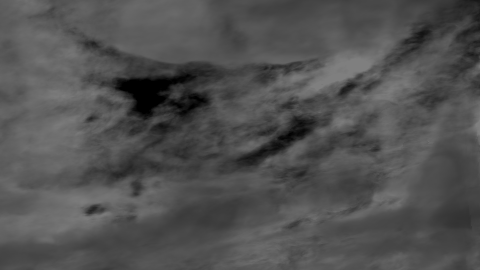}
\end{subfigure}
\begin{subfigure}{.33\linewidth}
    \includegraphics[\empty width=\linewidth]{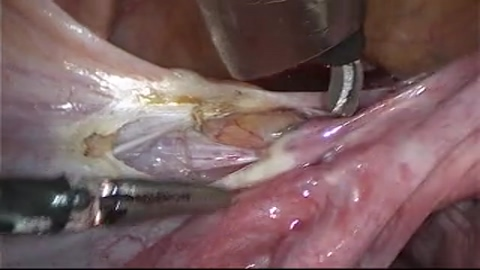}
\end{subfigure}


\begin{subfigure}{.33\linewidth}
    \includegraphics[\empty width=\linewidth]{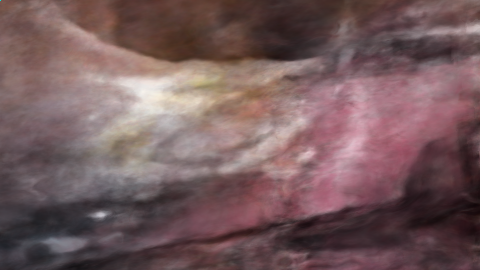}
\end{subfigure}
\begin{subfigure}{.33\linewidth}
    \includegraphics[\empty width=\linewidth]{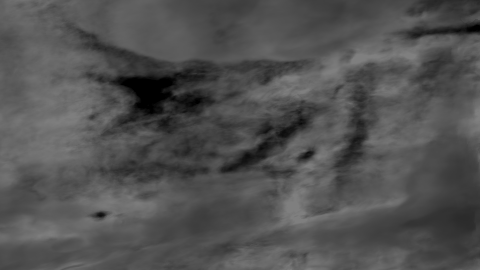}
\end{subfigure}
\begin{subfigure}{.33\linewidth}
    \includegraphics[\empty width=\linewidth]{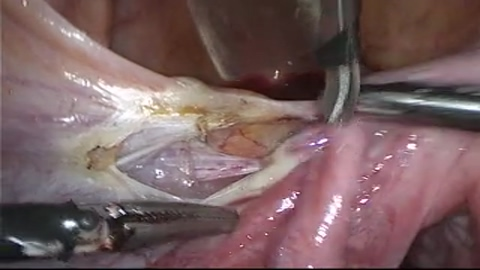}
\end{subfigure}

\caption[short]{\textbf{Qualitative results}: We present our qualitative results on the SurgicalActions160 dataset, specifically the \textbf{blunt dissection} action. \textit{Model reconstructions}, \textit{Depth predictions}, and \textit{Ground truth} (\textit{left} to \textit{right}) - Time-steps (\textit{top} to \textit{bottom})}
\label{fig:sa160-blunt-dissection}
\end{figure*}

\begin{figure*}
\centering


\begin{subfigure}{.33\linewidth}
    \includegraphics[\empty width=\linewidth]{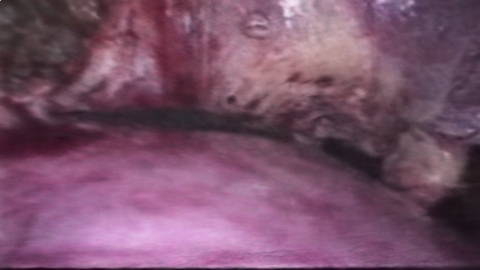}
\end{subfigure}
\begin{subfigure}{.33\linewidth}
    \includegraphics[\empty width=\linewidth]{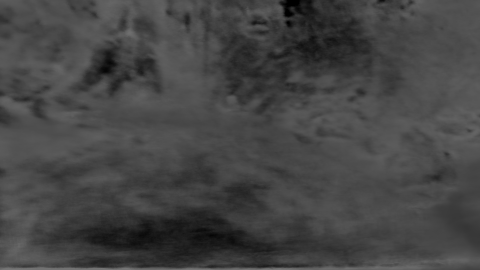}
\end{subfigure}
\begin{subfigure}{.33\linewidth}
    \includegraphics[\empty width=\linewidth]{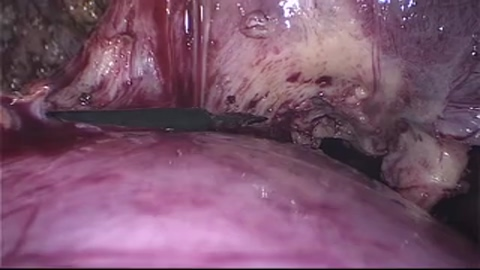}
\end{subfigure}


\begin{subfigure}{.33\linewidth}
    \includegraphics[\empty width=\linewidth]{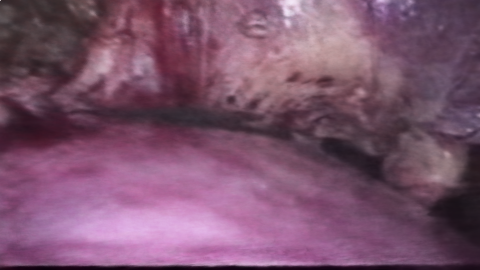}
\end{subfigure}
\begin{subfigure}{.33\linewidth}
    \includegraphics[\empty width=\linewidth]{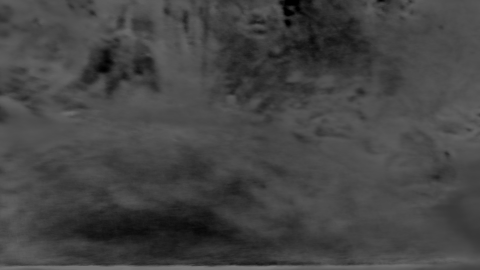}
\end{subfigure}
\begin{subfigure}{.33\linewidth}
    \includegraphics[\empty width=\linewidth]{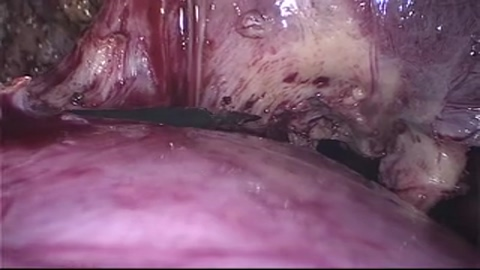}
\end{subfigure}


\begin{subfigure}{.33\linewidth}
    \includegraphics[\empty width=\linewidth]{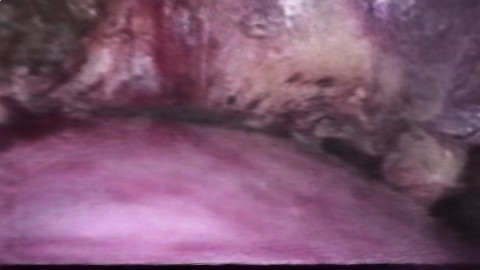}
\end{subfigure}
\begin{subfigure}{.33\linewidth}
    \includegraphics[\empty width=\linewidth]{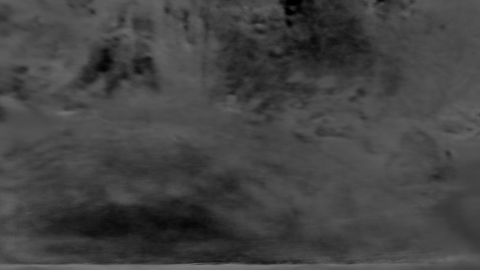}
\end{subfigure}
\begin{subfigure}{.33\linewidth}
    \includegraphics[\empty width=\linewidth]{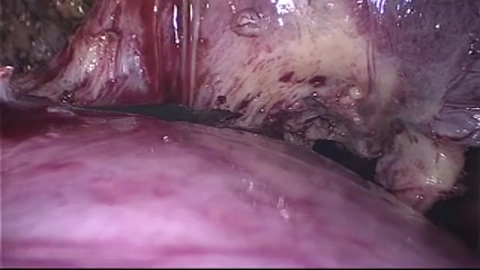}
\end{subfigure}


\begin{subfigure}{.33\linewidth}
    \includegraphics[\empty width=\linewidth]{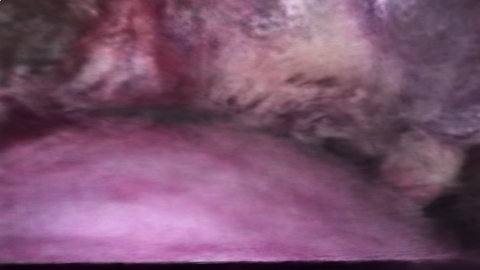}
\end{subfigure}
\begin{subfigure}{.33\linewidth}
    \includegraphics[\empty width=\linewidth]{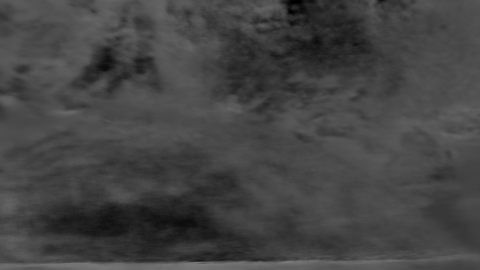}
\end{subfigure}
\begin{subfigure}{.33\linewidth}
    \includegraphics[\empty width=\linewidth]{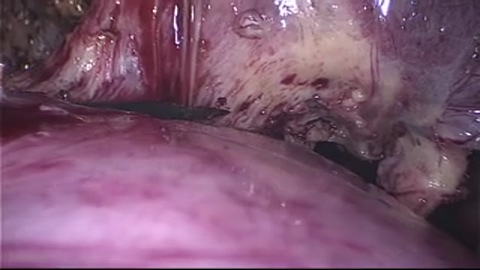}
\end{subfigure}


\begin{subfigure}{.33\linewidth}
    \includegraphics[\empty width=\linewidth]{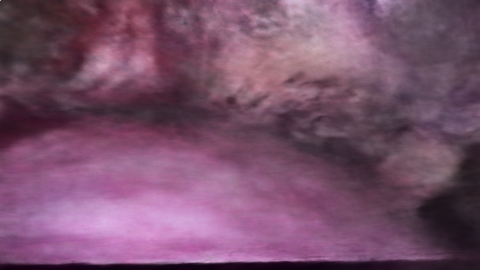}
\end{subfigure}
\begin{subfigure}{.33\linewidth}
    \includegraphics[\empty width=\linewidth]{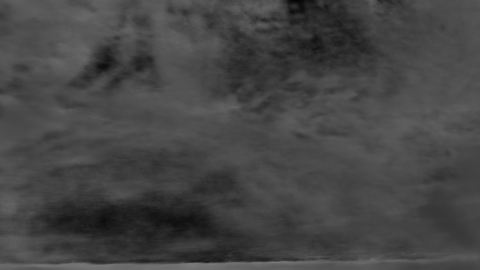}
\end{subfigure}
\begin{subfigure}{.33\linewidth}
    \includegraphics[\empty width=\linewidth]{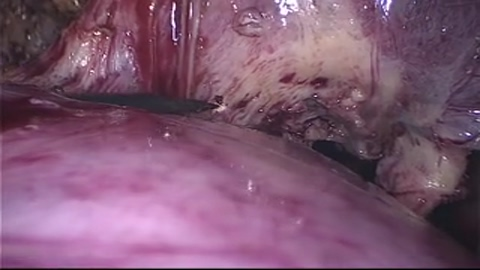}
\end{subfigure}


\begin{subfigure}{.33\linewidth}
    \includegraphics[\empty width=\linewidth]{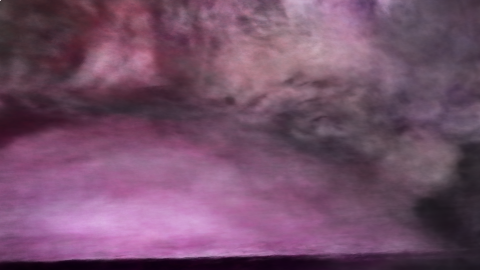}
\end{subfigure}
\begin{subfigure}{.33\linewidth}
    \includegraphics[\empty width=\linewidth]{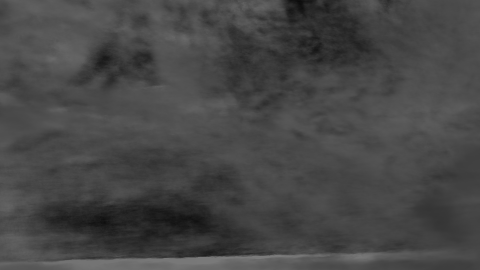}
\end{subfigure}
\begin{subfigure}{.33\linewidth}
    \includegraphics[\empty width=\linewidth]{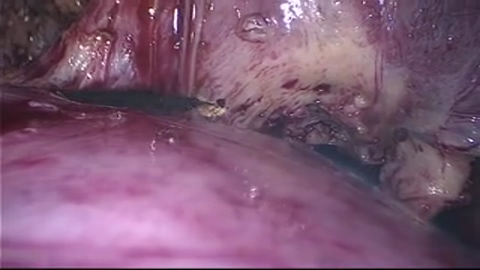}
\end{subfigure}

\caption[short]{\textbf{Qualitative results}: We present our qualitative results on the SurgicalActions160 dataset, specifically the \textbf{dissection thermal} action. \textit{Model reconstructions}, \textit{Depth predictions}, and \textit{Ground truth} (\textit{left} to \textit{right}) - Time-steps (\textit{top} to \textit{bottom})}
\label{fig:sa160-dissection-thermal}
\end{figure*}

\begin{figure*}
\centering


\begin{subfigure}{.33\linewidth}
    \includegraphics[\empty width=\linewidth]{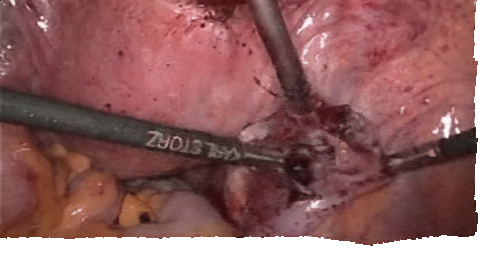}
\end{subfigure}
\begin{subfigure}{.33\linewidth}
    \includegraphics[\empty width=\linewidth]{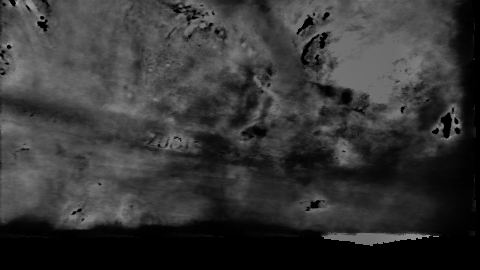}
\end{subfigure}
\begin{subfigure}{.33\linewidth}
    \includegraphics[\empty width=\linewidth]{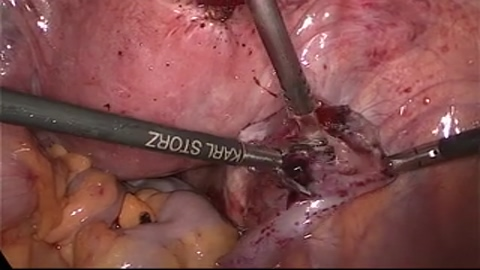}
\end{subfigure}


\begin{subfigure}{.33\linewidth}
    \includegraphics[\empty width=\linewidth]{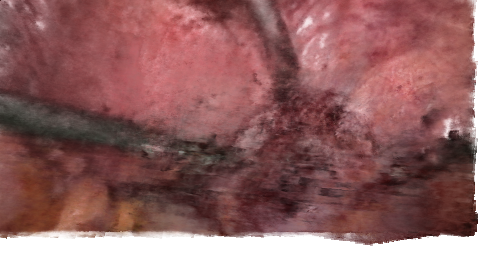}
\end{subfigure}
\begin{subfigure}{.33\linewidth}
    \includegraphics[\empty width=\linewidth]{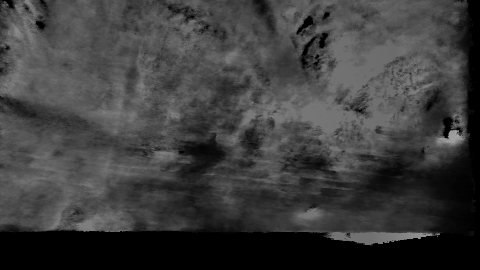}
\end{subfigure}
\begin{subfigure}{.33\linewidth}
    \includegraphics[\empty width=\linewidth]{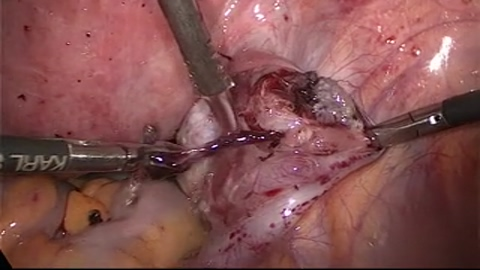}
\end{subfigure}


\begin{subfigure}{.33\linewidth}
    \includegraphics[\empty width=\linewidth]{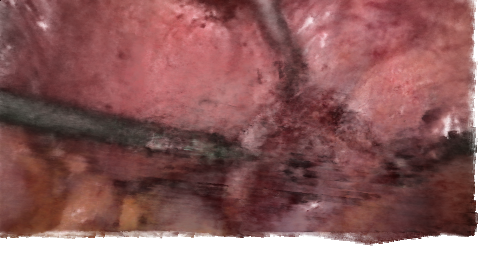}
\end{subfigure}
\begin{subfigure}{.33\linewidth}
    \includegraphics[\empty width=\linewidth]{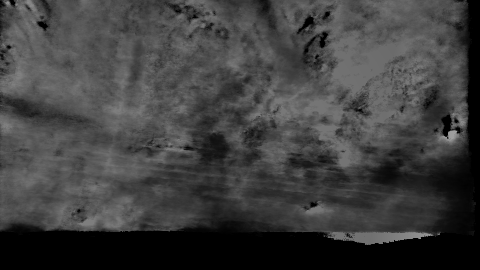}
\end{subfigure}
\begin{subfigure}{.33\linewidth}
    \includegraphics[\empty width=\linewidth]{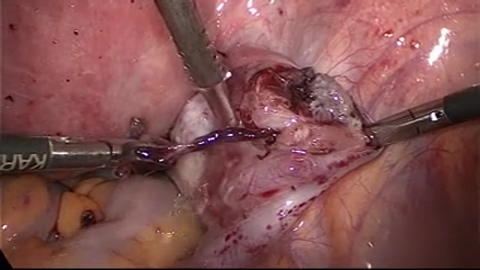}
\end{subfigure}


\begin{subfigure}{.33\linewidth}
    \includegraphics[\empty width=\linewidth]{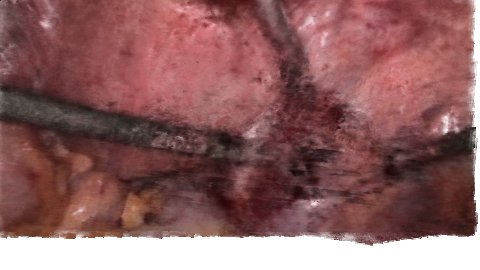}
\end{subfigure}
\begin{subfigure}{.33\linewidth}
    \includegraphics[\empty width=\linewidth]{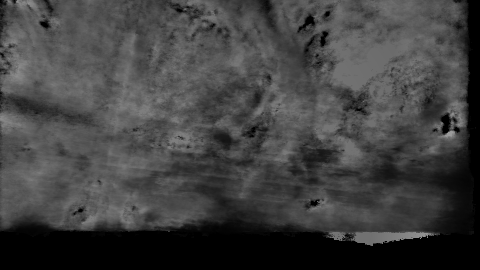}
\end{subfigure}
\begin{subfigure}{.33\linewidth}
    \includegraphics[\empty width=\linewidth]{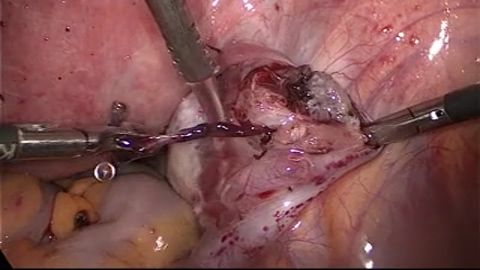}
\end{subfigure}


\begin{subfigure}{.33\linewidth}
    \includegraphics[\empty width=\linewidth]{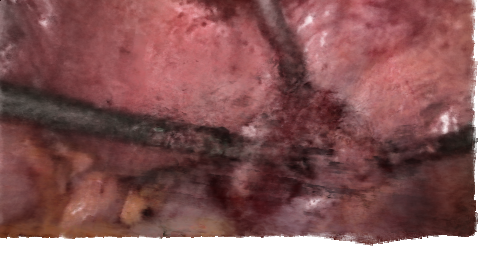}
\end{subfigure}
\begin{subfigure}{.33\linewidth}
    \includegraphics[\empty width=\linewidth]{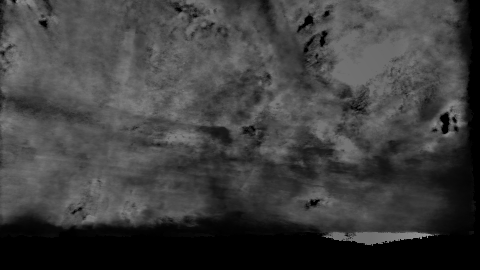}
\end{subfigure}
\begin{subfigure}{.33\linewidth}
    \includegraphics[\empty width=\linewidth]{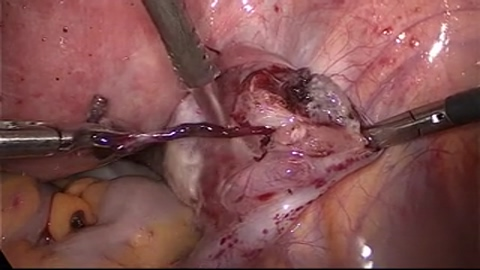}
\end{subfigure}

\caption[short]{\textbf{Qualitative results}: We present our qualitative results on the SurgicalActions160 dataset, specifically the \textbf{Irrigation} action. \textit{Model reconstructions}, \textit{Depth predictions}, and \textit{Ground truth} (\textit{left} to \textit{right}) - Time-steps (\textit{top} to \textit{bottom})}
\label{fig:sa160-irrigation}
\end{figure*}

\begin{figure*}
\centering


\begin{subfigure}{.33\linewidth}
    \includegraphics[\empty width=\linewidth]{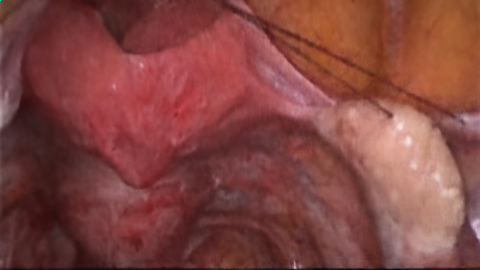}
\end{subfigure}
\begin{subfigure}{.33\linewidth}
    \includegraphics[\empty width=\linewidth]{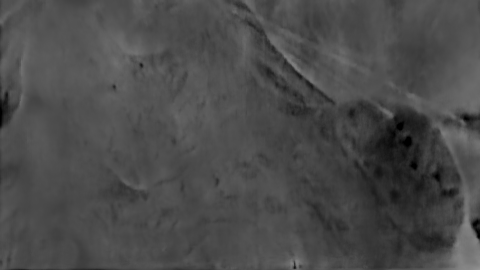}
\end{subfigure}
\begin{subfigure}{.33\linewidth}
    \includegraphics[\empty width=\linewidth]{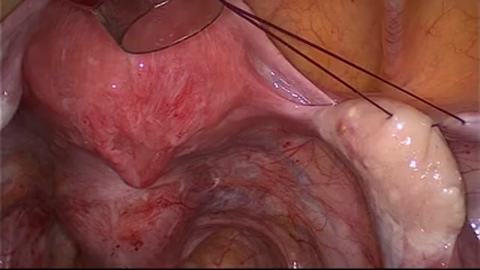}
\end{subfigure}


\begin{subfigure}{.33\linewidth}
    \includegraphics[\empty width=\linewidth]{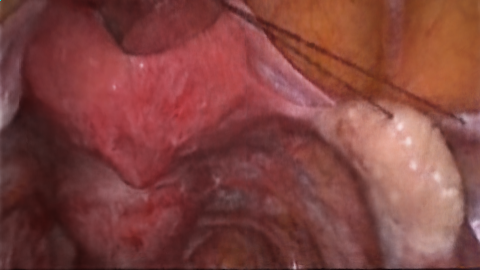}
\end{subfigure}
\begin{subfigure}{.33\linewidth}
    \includegraphics[\empty width=\linewidth]{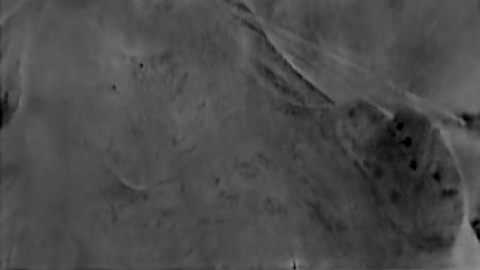}
\end{subfigure}
\begin{subfigure}{.33\linewidth}
    \includegraphics[\empty width=\linewidth]{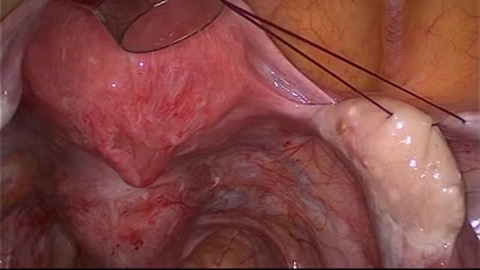}
\end{subfigure}


\begin{subfigure}{.33\linewidth}
    \includegraphics[\empty width=\linewidth]{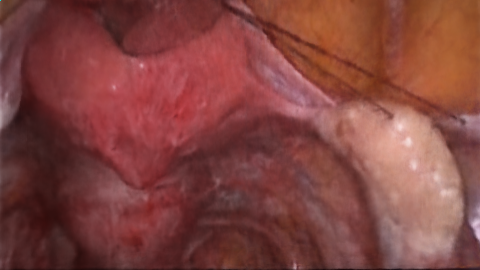}
\end{subfigure}
\begin{subfigure}{.33\linewidth}
    \includegraphics[\empty width=\linewidth]{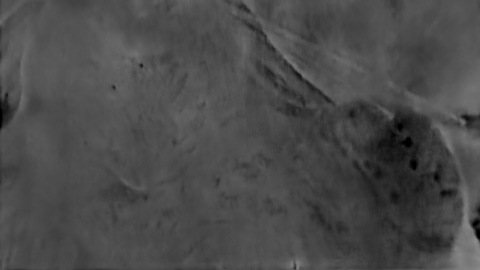}
\end{subfigure}
\begin{subfigure}{.33\linewidth}
    \includegraphics[\empty width=\linewidth]{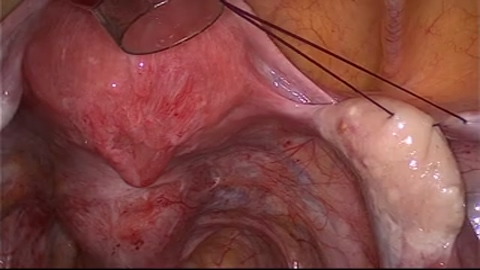}
\end{subfigure}


\begin{subfigure}{.33\linewidth}
    \includegraphics[\empty width=\linewidth]{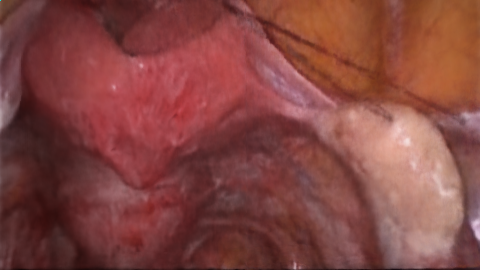}
\end{subfigure}
\begin{subfigure}{.33\linewidth}
    \includegraphics[\empty width=\linewidth]{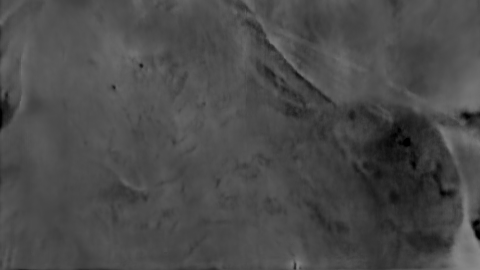}
\end{subfigure}
\begin{subfigure}{.33\linewidth}
    \includegraphics[\empty width=\linewidth]{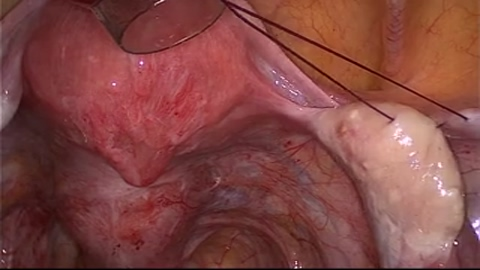}
\end{subfigure}


\begin{subfigure}{.33\linewidth}
    \includegraphics[\empty width=\linewidth]{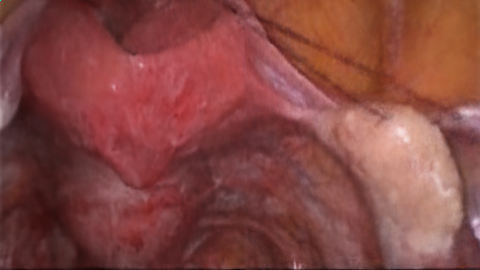}
\end{subfigure}
\begin{subfigure}{.33\linewidth}
    \includegraphics[\empty width=\linewidth]{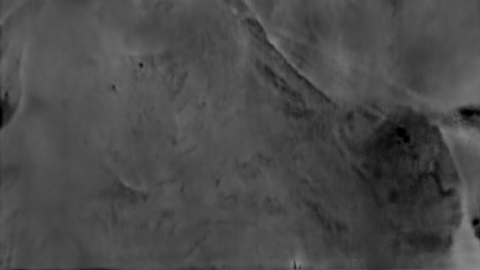}
\end{subfigure}
\begin{subfigure}{.33\linewidth}
    \includegraphics[\empty width=\linewidth]{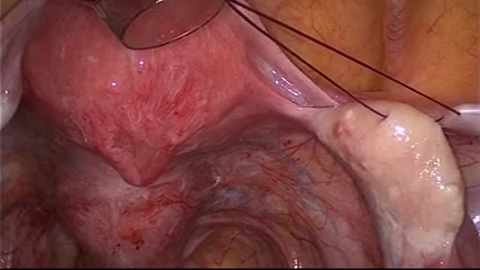}
\end{subfigure}


\begin{subfigure}{.33\linewidth}
    \includegraphics[\empty width=\linewidth]{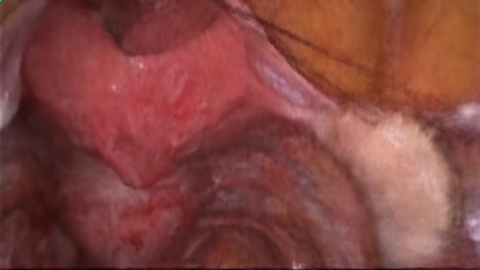}
\end{subfigure}
\begin{subfigure}{.33\linewidth}
    \includegraphics[\empty width=\linewidth]{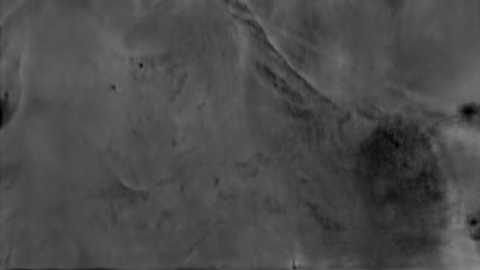}
\end{subfigure}
\begin{subfigure}{.33\linewidth}
    \includegraphics[\empty width=\linewidth]{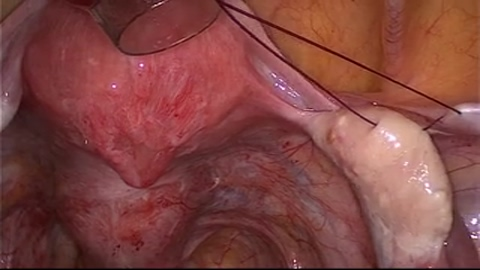}
\end{subfigure}

\caption[short]{\textbf{Qualitative results}: We present our qualitative results on the SurgicalActions160 dataset, specifically the \textbf{knot-pushing} action. \textit{Model reconstructions}, \textit{Depth predictions}, and \textit{Ground truth} (\textit{left} to \textit{right}) - Time-steps (\textit{top} to \textit{bottom})}
\label{fig:sa160-knot-pushing}
\end{figure*}

\begin{figure*}
\centering


\begin{subfigure}{.33\linewidth}
    \includegraphics[\empty width=\linewidth]{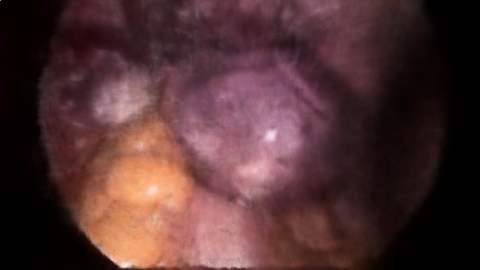}
\end{subfigure}
\begin{subfigure}{.33\linewidth}
    \includegraphics[\empty width=\linewidth]{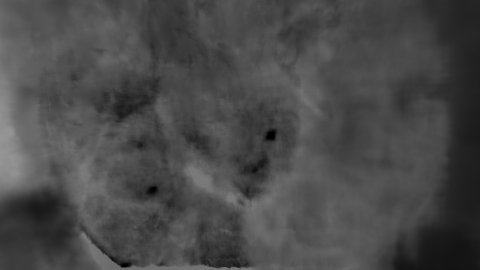}
\end{subfigure}
\begin{subfigure}{.33\linewidth}
    \includegraphics[\empty width=\linewidth]{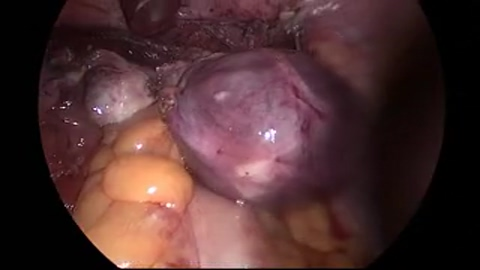}
\end{subfigure}


\begin{subfigure}{.33\linewidth}
    \includegraphics[\empty width=\linewidth]{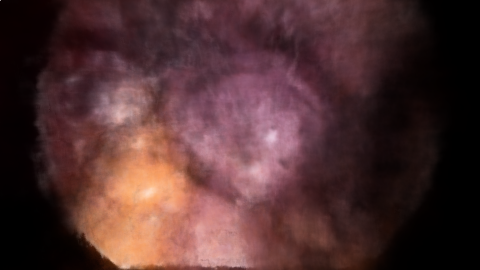}
\end{subfigure}
\begin{subfigure}{.33\linewidth}
    \includegraphics[\empty width=\linewidth]{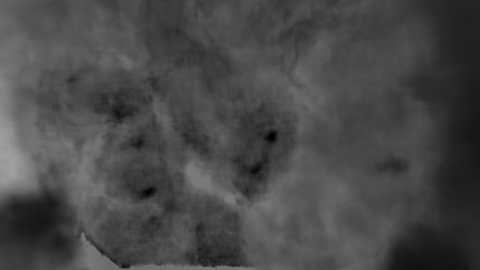}
\end{subfigure}
\begin{subfigure}{.33\linewidth}
    \includegraphics[\empty width=\linewidth]{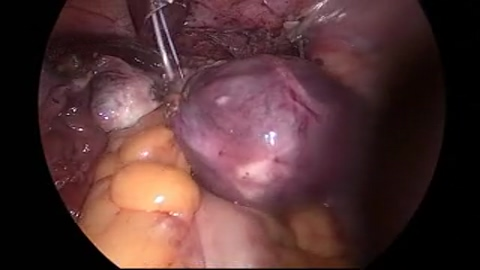}
\end{subfigure}


\begin{subfigure}{.33\linewidth}
    \includegraphics[\empty width=\linewidth]{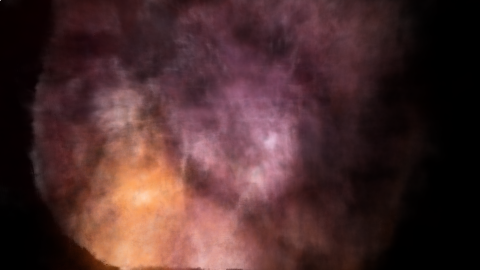}
\end{subfigure}
\begin{subfigure}{.33\linewidth}
    \includegraphics[\empty width=\linewidth]{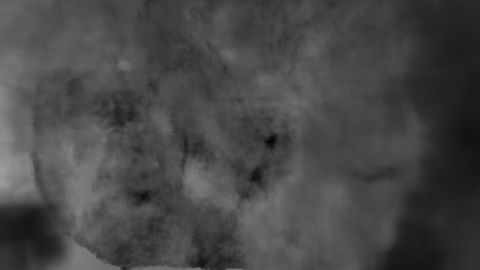}
\end{subfigure}
\begin{subfigure}{.33\linewidth}
    \includegraphics[\empty width=\linewidth]{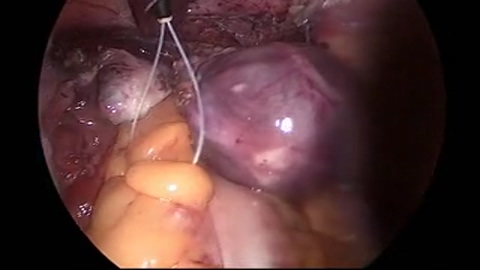}
\end{subfigure}


\begin{subfigure}{.33\linewidth}
    \includegraphics[\empty width=\linewidth]{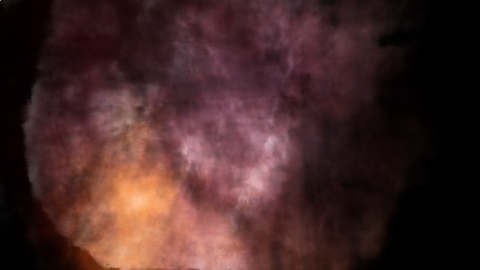}
\end{subfigure}
\begin{subfigure}{.33\linewidth}
    \includegraphics[\empty width=\linewidth]{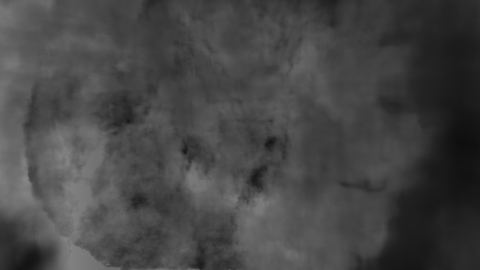}
\end{subfigure}
\begin{subfigure}{.33\linewidth}
    \includegraphics[\empty width=\linewidth]{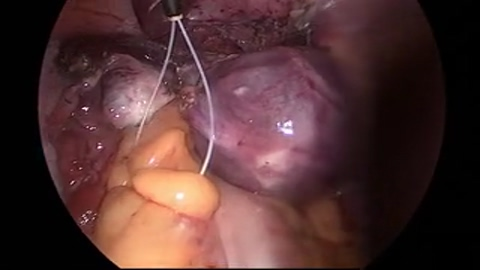}
\end{subfigure}


\begin{subfigure}{.33\linewidth}
    \includegraphics[\empty width=\linewidth]{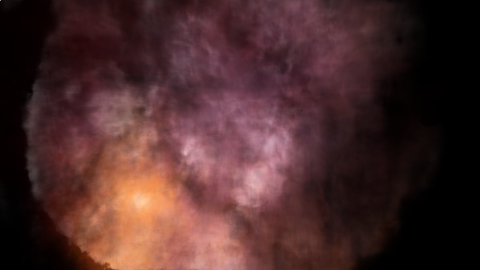}
\end{subfigure}
\begin{subfigure}{.33\linewidth}
    \includegraphics[\empty width=\linewidth]{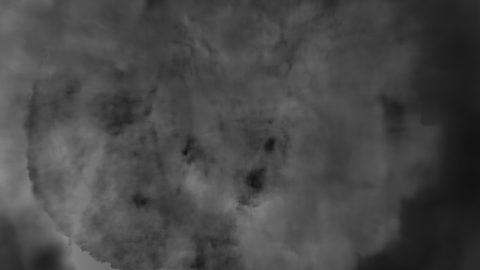}
\end{subfigure}
\begin{subfigure}{.33\linewidth}
    \includegraphics[\empty width=\linewidth]{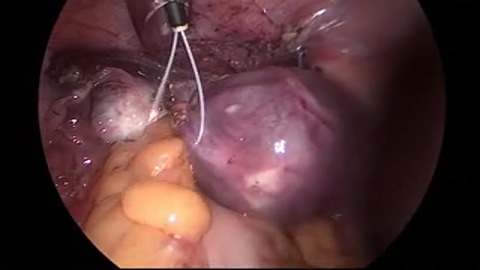}
\end{subfigure}


\begin{subfigure}{.33\linewidth}
    \includegraphics[\empty width=\linewidth]{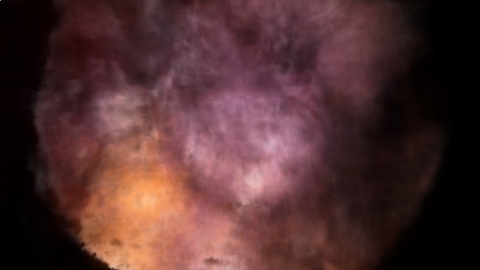}
\end{subfigure}
\begin{subfigure}{.33\linewidth}
    \includegraphics[\empty width=\linewidth]{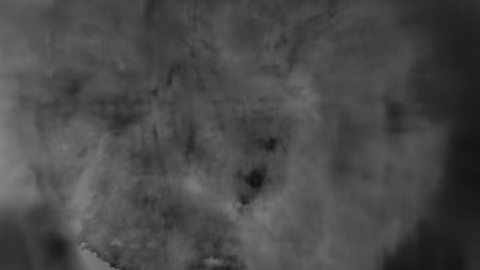}
\end{subfigure}
\begin{subfigure}{.33\linewidth}
    \includegraphics[\empty width=\linewidth]{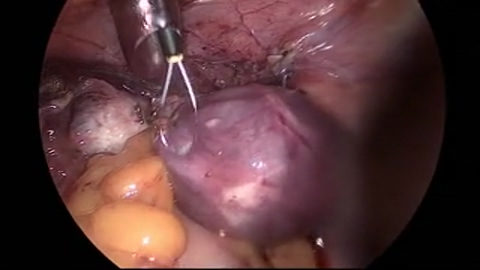}
\end{subfigure}

\caption[short]{\textbf{Qualitative results}: We present our qualitative results on the SurgicalActions160 dataset, specifically the \textbf{sling-in} action. \textit{Model reconstructions}, \textit{Depth predictions}, and \textit{Ground truth} (\textit{left} to \textit{right}) - Time-steps (\textit{top} to \textit{bottom})}
\label{fig:sa160-sling-in}
\end{figure*}

\begin{figure*}
\centering


\begin{subfigure}{.33\linewidth}
    \includegraphics[\empty width=\linewidth]{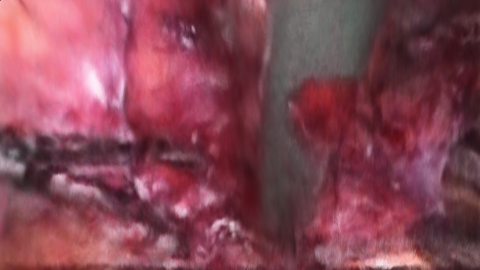}
\end{subfigure}
\begin{subfigure}{.33\linewidth}
    \includegraphics[\empty width=\linewidth]{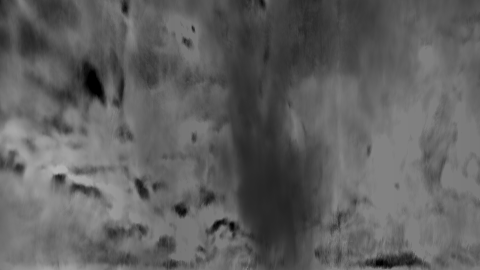}
\end{subfigure}
\begin{subfigure}{.33\linewidth}
    \includegraphics[\empty width=\linewidth]{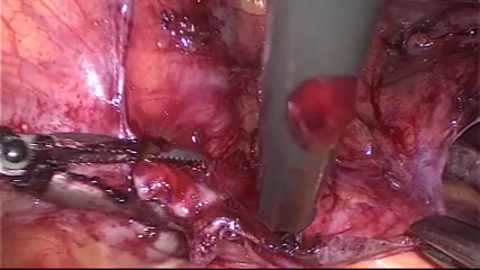}
\end{subfigure}


\begin{subfigure}{.33\linewidth}
    \includegraphics[\empty width=\linewidth]{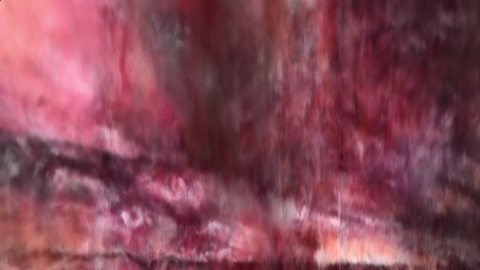}
\end{subfigure}
\begin{subfigure}{.33\linewidth}
    \includegraphics[\empty width=\linewidth]{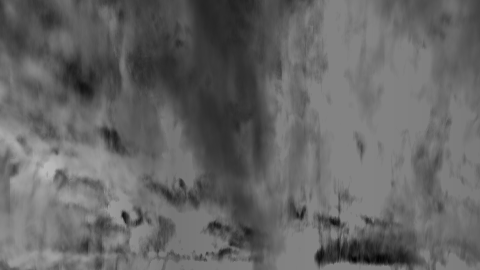}
\end{subfigure}
\begin{subfigure}{.33\linewidth}
    \includegraphics[\empty width=\linewidth]{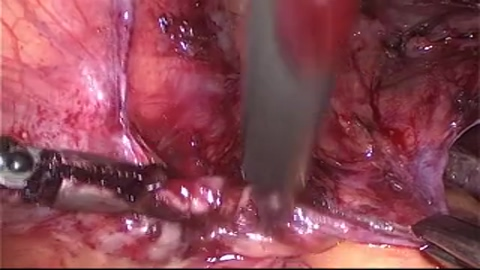}
\end{subfigure}


\begin{subfigure}{.33\linewidth}
    \includegraphics[\empty width=\linewidth]{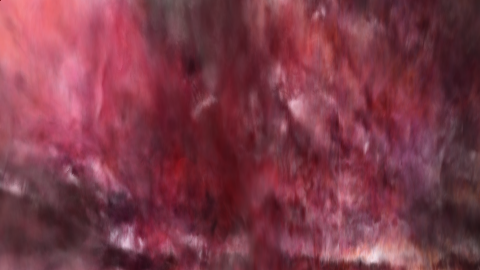}
\end{subfigure}
\begin{subfigure}{.33\linewidth}
    \includegraphics[\empty width=\linewidth]{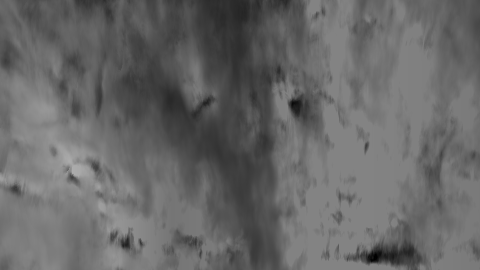}
\end{subfigure}
\begin{subfigure}{.33\linewidth}
    \includegraphics[\empty width=\linewidth]{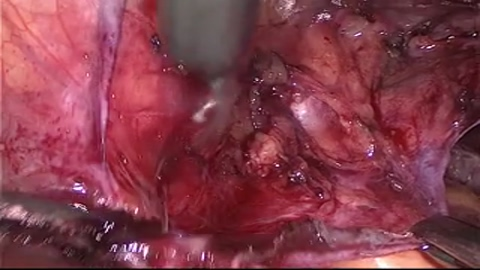}
\end{subfigure}


\begin{subfigure}{.33\linewidth}
    \includegraphics[\empty width=\linewidth]{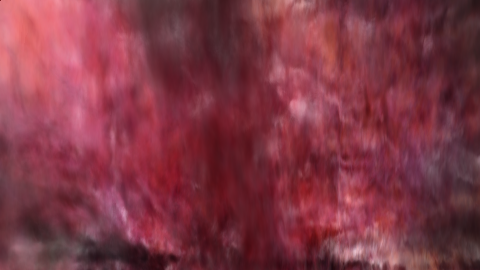}
\end{subfigure}
\begin{subfigure}{.33\linewidth}
    \includegraphics[\empty width=\linewidth]{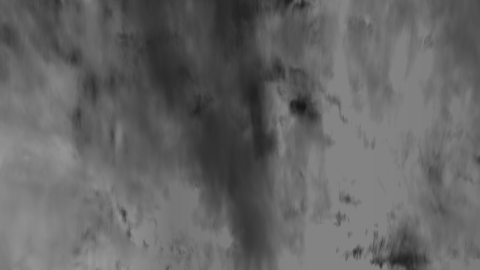}
\end{subfigure}
\begin{subfigure}{.33\linewidth}
    \includegraphics[\empty width=\linewidth]{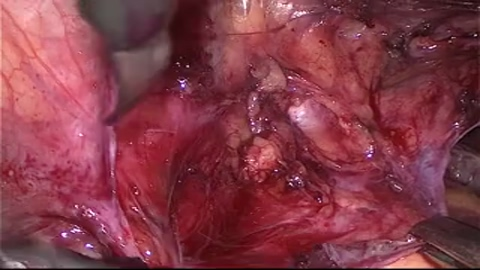}
\end{subfigure}


\begin{subfigure}{.33\linewidth}
    \includegraphics[\empty width=\linewidth]{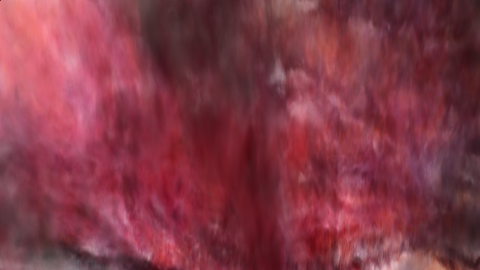}
\end{subfigure}
\begin{subfigure}{.33\linewidth}
    \includegraphics[\empty width=\linewidth]{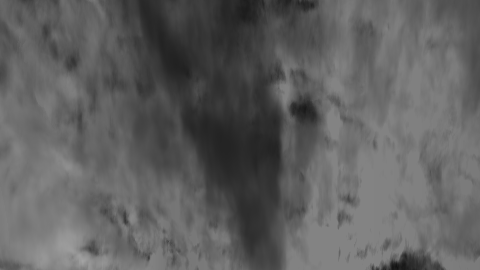}
\end{subfigure}
\begin{subfigure}{.33\linewidth}
    \includegraphics[\empty width=\linewidth]{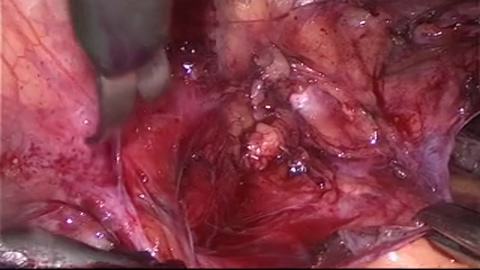}
\end{subfigure}


\begin{subfigure}{.33\linewidth}
    \includegraphics[\empty width=\linewidth]{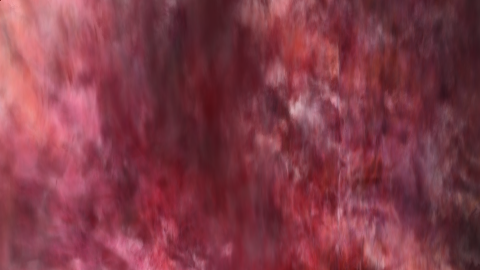}
\end{subfigure}
\begin{subfigure}{.33\linewidth}
    \includegraphics[\empty width=\linewidth]{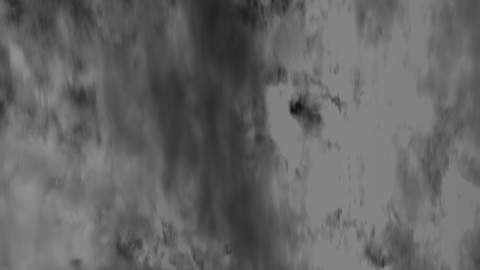}
\end{subfigure}
\begin{subfigure}{.33\linewidth}
    \includegraphics[\empty width=\linewidth]{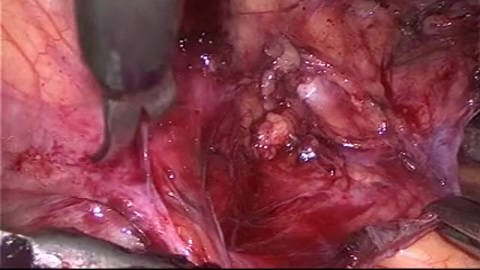}
\end{subfigure}

\caption[short]{\textbf{Qualitative results}: We present our qualitative results on the SurgicalActions160 dataset, specifically the \textbf{cutting} action. \textit{Model reconstructions}, \textit{Depth predictions}, and \textit{Ground truth} (\textit{left} to \textit{right}) - Time-steps (\textit{top} to \textit{bottom})}
\label{fig:sa160-cutting}
\end{figure*}

The Surgical Actions dataset consists of 16 different surgical actions and 10 short samples of each.
We select the following: \textit{Cutting} (Figure \ref{fig:sa160-cutting}), \textit{Blunt dissection} (Figure \ref{fig:sa160-blunt-dissection}), \textit{Dissection thermal} (Figure \ref{fig:sa160-dissection-thermal}), \textit{Irrigation} (Figure \ref{fig:sa160-irrigation}), \textit{Knot-pushing} (Figure \ref{fig:sa160-knot-pushing}), and \textit{Sling-in} (Figure \ref{fig:sa160-sling-in}).

\subsection{Cholec80}
\begin{figure*}
\centering


\begin{subfigure}{.33\linewidth}
    \includegraphics[\empty width=\linewidth]{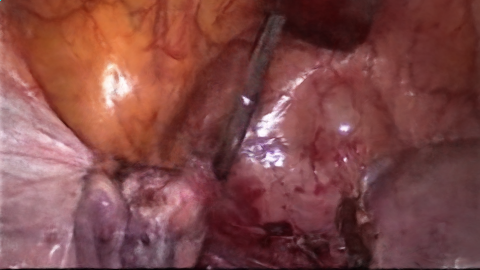}
\end{subfigure}
\begin{subfigure}{.33\linewidth}
    \includegraphics[\empty width=\linewidth]{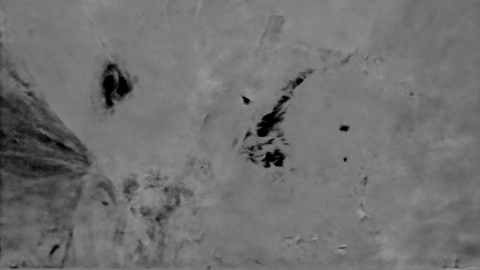}
\end{subfigure}
\begin{subfigure}{.33\linewidth}
    \includegraphics[\empty width=\linewidth]{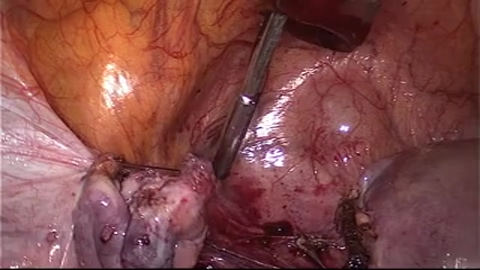}
\end{subfigure}


\begin{subfigure}{.33\linewidth}
    \includegraphics[\empty width=\linewidth]{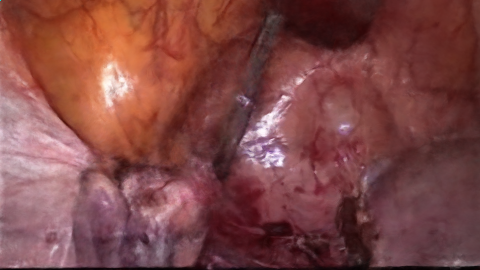}
\end{subfigure}
\begin{subfigure}{.33\linewidth}
    \includegraphics[\empty width=\linewidth]{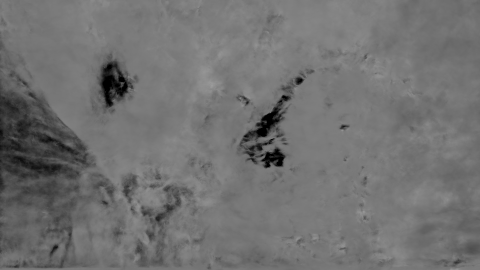}
\end{subfigure}
\begin{subfigure}{.33\linewidth}
    \includegraphics[\empty width=\linewidth]{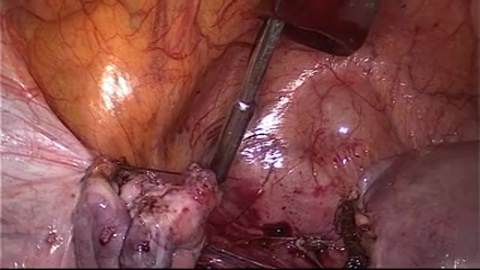}
\end{subfigure}


\begin{subfigure}{.33\linewidth}
    \includegraphics[\empty width=\linewidth]{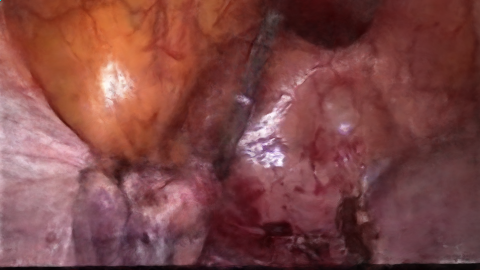}
\end{subfigure}
\begin{subfigure}{.33\linewidth}
    \includegraphics[\empty width=\linewidth]{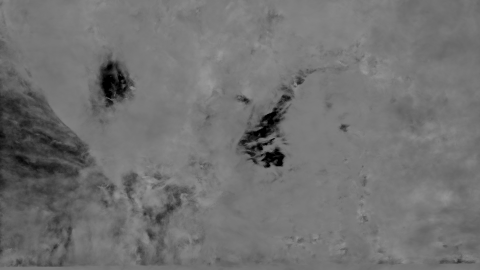}
\end{subfigure}
\begin{subfigure}{.33\linewidth}
    \includegraphics[\empty width=\linewidth]{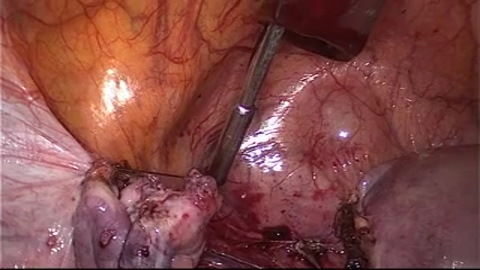}
\end{subfigure}


\begin{subfigure}{.33\linewidth}
    \includegraphics[\empty width=\linewidth]{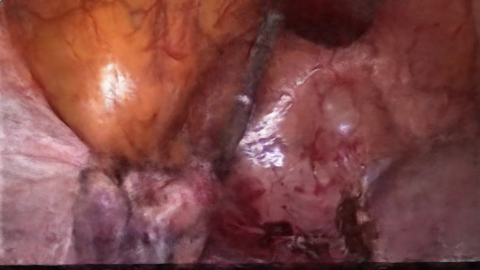}
\end{subfigure}
\begin{subfigure}{.33\linewidth}
    \includegraphics[\empty width=\linewidth]{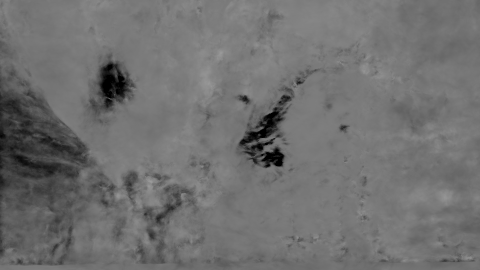}
\end{subfigure}
\begin{subfigure}{.33\linewidth}
    \includegraphics[\empty width=\linewidth]{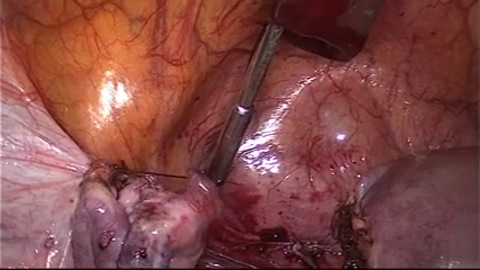}
\end{subfigure}


\begin{subfigure}{.33\linewidth}
    \includegraphics[\empty width=\linewidth]{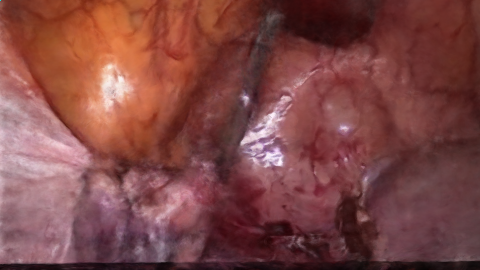}
\end{subfigure}
\begin{subfigure}{.33\linewidth}
    \includegraphics[\empty width=\linewidth]{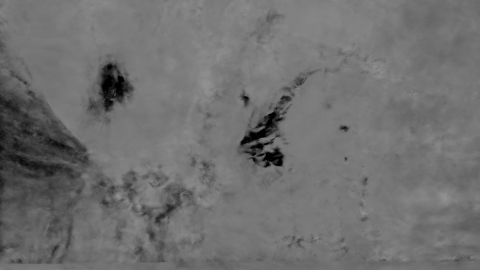}
\end{subfigure}
\begin{subfigure}{.33\linewidth}
    \includegraphics[\empty width=\linewidth]{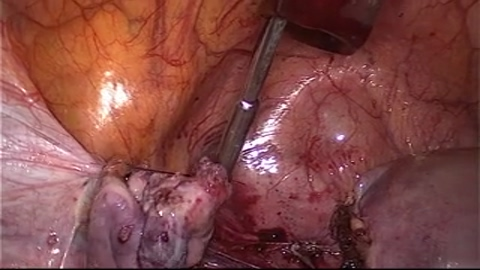}
\end{subfigure}

\caption[short]{\textbf{Qualitative results}: We present our qualitative results on the SurgicalActions160 dataset. \textit{Model reconstructions}, \textit{Depth predictions}, and \textit{Ground truth} (\textit{left} to \textit{right}) - Time-steps (\textit{top} to \textit{bottom})}
\label{fig:cholec80}
\end{figure*}

The inclusion of the Cholec80 dataset is important as it provides an illustration of possible extensions of this method to the extremely challenging datatype - monocular surgical data. We choose only a very short clip extracted from the dataset and show the resulting outputs (Figure \ref{fig:cholec80}). For reference, we sub-sample the clips so as to not have to process and thus reconstruct every frame.

\end{document}